\documentclass[letterpaper, 10 pt, journal, twoside]{IEEEtran}
\usepackage{array,multirow,graphics}
\usepackage{graphicx}
\usepackage{times}
\usepackage{amsmath}
\usepackage{amssymb}
\usepackage{xcolor}
\usepackage{textcomp}
\usepackage{siunitx}
\usepackage{balance}
\usepackage[hidelinks]{hyperref}
\usepackage{url}
\usepackage{fancyhdr}

\def\secref#1{Sec.~\ref{#1}}
\def\figref#1{Fig.~\ref{#1}}
\def\tabref#1{Tab.~\ref{#1}}
\def\eqref#1{Eq.~(\ref{#1})}

\let\oldmaketitle\maketitle
 
\renewcommand{\maketitle}{%
  \oldmaketitle
  \thispagestyle{fancy}
}

\begin{document}

\title{Building an Aerial-Ground Robotics System \\for Precision Farming: An Adaptable Solution}
%
%
\author{Alberto Pretto$^{1,9,*}$, St{\'e}phanie Aravecchia$^{2}$, Wolfram Burgard$^{3}$, Nived Chebrolu$^{4}$, Christian Dornhege$^{3}$, Tillmann Falck$^{5}$, Freya Fleckenstein$^{3}$, Alessandra Fontenla$^{6}$, Marco Imperoli$^{1}$, Raghav Khanna$^{7}$, Frank Liebisch$^{8}$, Philipp Lottes$^{4}$, Andres Milioto$^{4}$, Daniele Nardi$^{1}$, Sandro Nardi$^{6}$, Johannes Pfeifer$^{8}$, Marija Popovi{\'c}$^{7,10}$, Ciro Potena$^{1}$, C{\'e}dric Pradalier$^{2}$, Elisa Rothacker-Feder$^{5}$, Inkyu Sa$^{7,11}$,  Alexander Schaefer$^{3}$, Roland Siegwart$^{7,\dagger}$, Cyrill Stachniss$^{4}$, Achim Walter$^{8}$, Wera Winterhalter$^{3}$,  Xiaolong Wu$^{2}$ and Juan Nieto$^{7}$
\thanks{This work was supported by the EC under Grant H2020-ICT-644227-Flourish and by the Swiss State Secretariat for Education, Research and Innovation under contract number 15.0029.}
\thanks{$^{1}$Department of Computer, Control, and Management Engineering, Sapienza University of Rome, Rome, Italy. $^{2}$International Research Lab 2958 Georgia Tech-CNRS, Metz, France. $^{3}$Department of Computer Science, University of Freiburg, Germany. Wolfram Burgard is also with the Toyota Research Institute, Los Altos, USA. $^{4}$ Photogrammetry \& Robotics Lab, University of Bonn, Germany. $^{5}$Robert Bosch GmbH, Corporate Research, Renningen, Germany. $^{6}$Agency for Agro-food Sector
Services of the Marche Region (ASSAM), Osimo, Italy. $^{7}$Autonomous Systems Lab., Department of Mechanical and Process Engineering, ETH Zurich, Z{\"u}rich, Switzerland. $^{8}$Department of Environmental Systems Science, Institute of Agricultural Sciences, ETH Zurich, Z{\"u}rich, Switzerland.
$^{9}$IT+Robotics Srl, Padova, Italy,
$^{10}$Smart Robotics Lab., Department of Computing, Imperial College London, London, UK.
$^{11}$CSIRO Data61, Australia.
$^{*}$Corresponding author.$^{\dagger}$Flourish project coordinator.}}

\maketitle              
%
\begin{abstract}
The application of autonomous robots in agriculture is gaining increasing popularity 
thanks to the high impact it may have on food security, sustainability, resource use efficiency, reduction of chemical treatments, 
and the optimization of human effort and yield. 
With this vision, the Flourish research project 
aimed to develop an adaptable robotic solution for
precision farming that combines the aerial survey capabilities of small autonomous unmanned aerial vehicles (UAVs) with 
targeted intervention performed by multi-purpose
unmanned ground vehicles (UGVs). This paper presents an overview of the 
scientific and technological advances and outcomes obtained in the 
project. 
We introduce multi-spectral perception algorithms and
aerial and ground-based systems developed for monitoring
crop density, weed pressure, crop nitrogen nutrition status, and to accurately classify and locate weeds. We then introduce
the navigation and mapping systems 
tailored to our robots in the agricultural environment, as well as the modules for collaborative mapping.
We finally present the ground
intervention hardware, software solutions, and interfaces we implemented and tested in different field
conditions and with different crops. We describe 
a real use case in which a UAV collaborates with a
UGV to monitor the field and to perform selective spraying
without human intervention.

\end{abstract}


\section{Introduction}

\begin{figure}[t!]
   \centering
   \includegraphics[width=\linewidth]{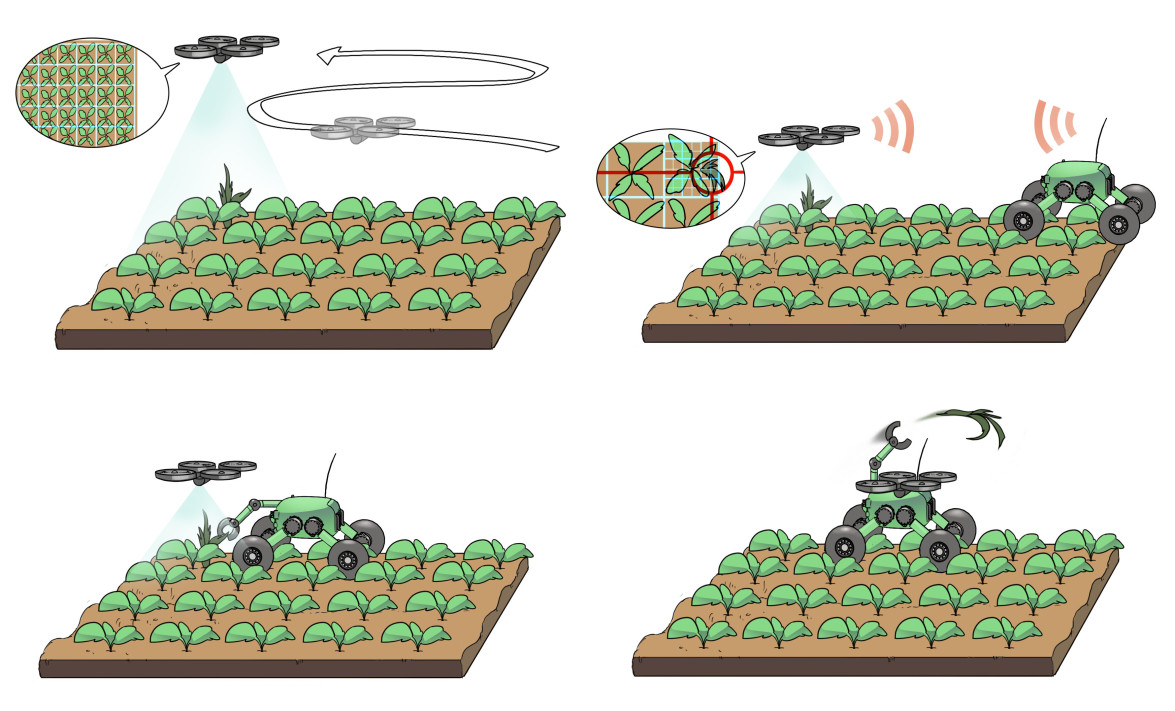}
   \caption{A conceptual overview of the Flourish project. A UAV continuously surveys a field over the growing season (top left), collecting data about crop density and weed pressure (top right) and coordinating and sharing information with a UGV (bottom left) that is used for targeted intervention and data analysis (bottom right). The gathered and merged information is then delivered to farm operators for high-level decision making.}
   \label{fig:concept}
\end{figure}

Collaborative aerial- and ground-based robotic systems offer significant benefits to many practical applications, as they can merge the advantages of multiple heterogeneous platforms.
This is especially useful 
for precision agriculture scenarios, where areas of interest are usually vast. For example, an Unmanned Aerial Vehicle (UAV) allows for rapid inspections of large areas, e.g., mapping weed distributions or crop nutrition status indicators.
This information can then be shared with an Unmanned Ground Vehicle (UGV), which can perform targeted actions, e.g., selective weed treatment or fertilizer applications, on required areas,
with relatively high operating times and payload capacities.
One of the main objectives of the Flourish project~\cite{flourish_project} (see \figref{fig:concept}) was to 
exploit this combined workflow in an autonomous robotic system for precision agriculture that achieves high yields while minimizing on-field chemical applications via targeted intervention.

This paper presents an overview of the scientific and technical outcomes obtained within the Flourish project, providing insights and practical details on the lessons learned in several areas ranging from robot navigation, mapping and coordination, up to robot vision, multi-spectral data analysis, and phenotyping.

To develop the experimental robots, we built upon existing state-of-the-art aerial and ground platforms, extending them in various aspects (\secref{sec:platforms}), with the installation of a large number of specific sensors, built-in computing power, and modules for weed detection, tracking, and removal. We proposed several novel perception methods and algorithms designed to automatically perform cyclical and exhaustive in-field measurements and interpretations (\secref{sec:data_analysis}), such as inference of weed density from multi-spectral images, mapping and classification of crops and weeds, and computation of plant health indicators.
Robot positioning and cooperative environment modeling (\secref{sec:env_modeling}) have been addressed proposing novel algorithms that leverage specific field representations, or building upon existing methods tailored for the specificity of the environment, e.g., by integrating multi-spectral imaging in the UAV mapping algorithms and by exploiting specific environment priors into the UGV positioning algorithms and the temporal map registration algorithms.
We propose a novel mission planner that allows the UAV to adaptively map large areas while respecting battery constraints, while we addressed safe UGV navigation in a cultivated field by integrating accurate relative localization, crop row detection, and an ad-hoc controller (\secref{sec:planning_navigation}); the mission coordination has been assured by a lightweight task scheduler and communication framework. We finally present a use case of ground intervention in the field, by means of accurate weed tracking for precision tool placement, and the development of a selective spraying and mechanical weed treatment module that is suitable for commercialization (\secref{in-field_intervention}). Another important outcome of the project is the large amount of open-source software modules released and datasets generated, which we hope the community will benefit from (\secref{sec:datasets}).

\subsection{Robotics in Agriculture: An Overview of Recent Projects}

Robotic applications in agriculture have a significant potential to improve field monitoring and intervention procedures. However, these technologies are still in a development 
phase, with many possible uses yet to be explored.

A project similar to Flourish is RHEA~\cite{rhea_project}, that aims at diminishing the use of agricultural chemical inputs by 75\%, improving crop quality, 
human health and safety, and reducing production costs by means of sustainable crop management using a fleet of small, heterogeneous robots (ground and aerial) equipped with advanced sensors, enhanced endeffectors, and improved decision control algorithms. Likewise, the PANTHEON project~\cite{gasparri_h2020_2018} aims to develop a Supervisory Control And Data Acquisition (SCADA) system for
precision farming in hazelnut orchards using a team of aerial-ground robots.

Other recent projects dealing with the development of autonomous ground platforms are the GRAPE~\cite{astofli_grapes_2018} and the SWEEPER~\cite{sweeper_project} projects. The former aims at creating agricultural service companies and equipment providers to develop vineyard robots that can increase the cost effectiveness of their products as compared to traditional practices. In particular, the project addresses the market of instruments for biological control by developing the tools required to execute (semi) autonomous vineyard monitoring and farming tasks with UGVs and, therefore, reducing environmental impact with respect to traditional chemical control. The SWEEPER main objective is to put the first generation greenhouse harvesting robots onto the market.

UAVs are increasingly used in many agricultural robotics applications, e.g. for tree 3D reconstruction and canopy estimation~\cite{Dong2019}, fruit counting~\cite{Chen2017}, yield estimation~\cite{Ehsani2016}, and 
automated monitoring using light-weight devices~\cite{Das2015}.

On the industry side, several start-ups have been raised, and 
many more expect to be funded.
The major services provided are UGVs for weed removal~\cite{ecorobotix}\cite{blueriver}\cite{sagarobotics}, and in-season data analytics or early pest and disease detection from aerial or satellite imagery.

\section{Experimental Platforms}\label{sec:platforms}

\begin{figure*}[th!]
   \centering
   \begin{minipage}[b]{0.32\linewidth}
   \begin{minipage}[b]{.99\linewidth}
   \includegraphics[width=\columnwidth]{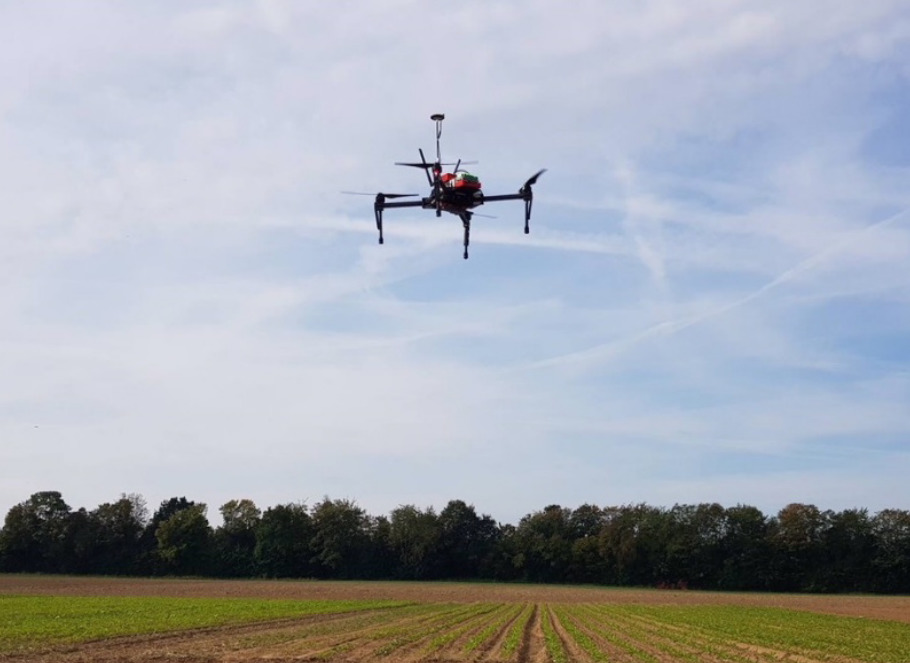}
   \end{minipage}\hfill
   \vspace{2.5mm}
   \begin{minipage}[b]{.99\linewidth}
   \includegraphics[width=\columnwidth,height=51mm]{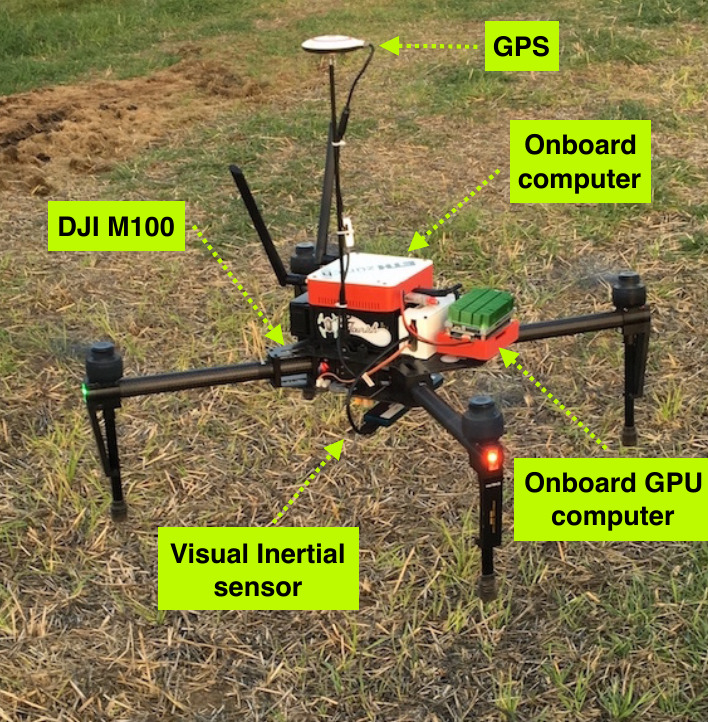}
   \end{minipage}\hfill
   \end{minipage}\hfill
   \begin{minipage}[b]{0.67\linewidth}
   \includegraphics[width=\columnwidth]{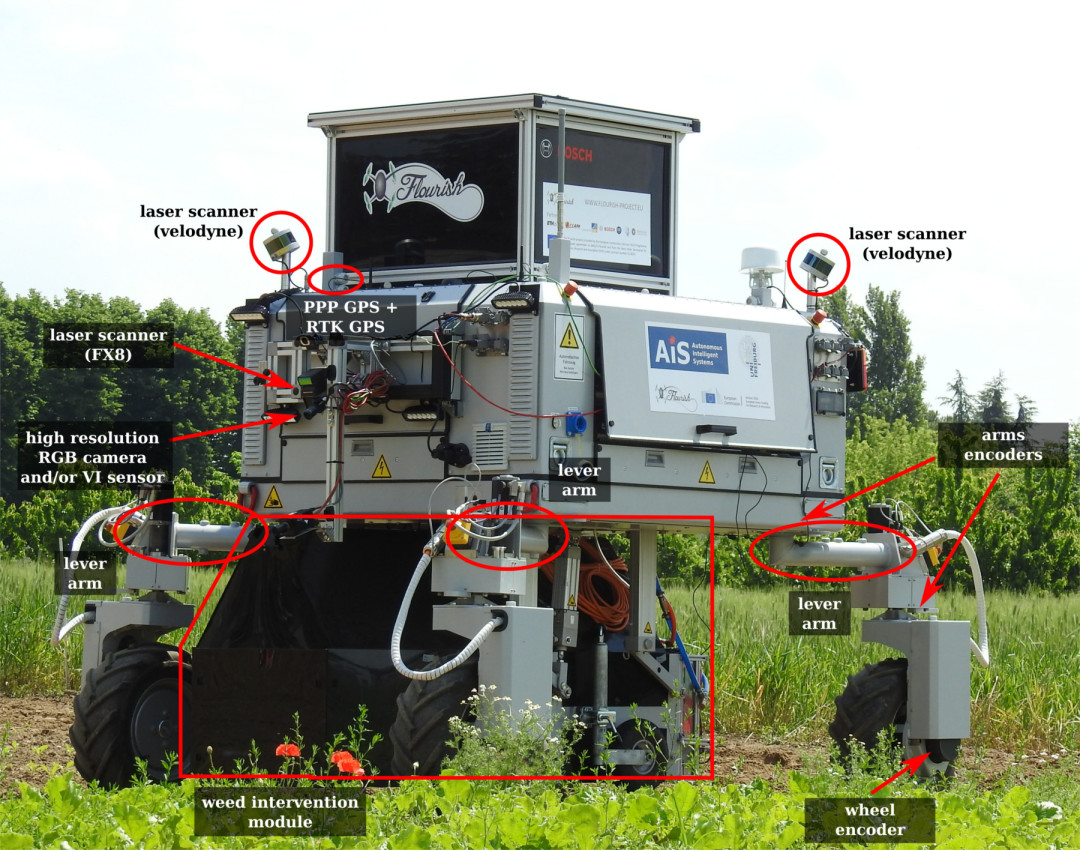}
   \end{minipage}\hfill
   \caption{The two main robots used in the experiments and demonstrations: a DJI Matrice 100 UAV multi-rotor performing an autonomous flight over a sugar beet field (top-left); the UAV with highlighted the installed sensors (bottom-left) and the Bosch BoniRob farming UGV (right).}
   \label{fig:uav_ugv}
\end{figure*}

The Flourish project exploited existing state-of-the-art farming and aerial robots, extending them in various aspects to improve both autonomous navigation and environment modeling capabilities, and to enable them to perform robust plant classification and/or selective weed removal operations.

\subsection{Multirotor Used in the Flourish Project}

The main UAV platform in the project is a fully sensorized DJI Matrice 100 (\figref{fig:uav_ugv}, left). The platform includes an Intel NUC i7 computer for on-board processing, a NVIDIA TX2 GPU for real-time weed detection, a GPS module, and a visual-inertial (VI) system for egomotion estimation. We employ a VI sensor developed at the Autonomous Systems Lab.~\cite{Nikolic2014}, and also tested and integrated a commercially available sensor, the Intel ZR300, for wider usage. 

\subsection{Ground Vehicle}
\label{sec:bonirob}

\subsubsection{The BoniRob Farming Robot}
The Bosch Deepfield Robotics BoniRob (\figref{fig:uav_ugv}, right) is a flexible research platform for agricultural robotics. 
Its four wheels can be independently rotated around the vertical axis, resulting in omnidirectional driving capabilities, and are mounted at the end of lever arms, letting the robot adjust its track width from \SI{1}{m} to \SI{2}{m}. 
In order to execute complex tasks, the BoniRob carries a multitude of sensors: GPS, RTK-GPS, a push-broom lidar, two omnidirectional lidars, RGB cameras, a VI system, hyperspectral cameras, wheel odometers, etc.
These sensors are directly connected to a set of on-board PCs that run the Robot Operating System (ROS) and communicate through an internal network.
The BoniRob's batteries are complemented by a backup generator that facilitates long-term field application. 

\subsubsection{Weed Intervention Module}\label{sec:weed_unit_sensors}
\begin{figure}[t!]
   \centering
     \includegraphics[width=\linewidth]{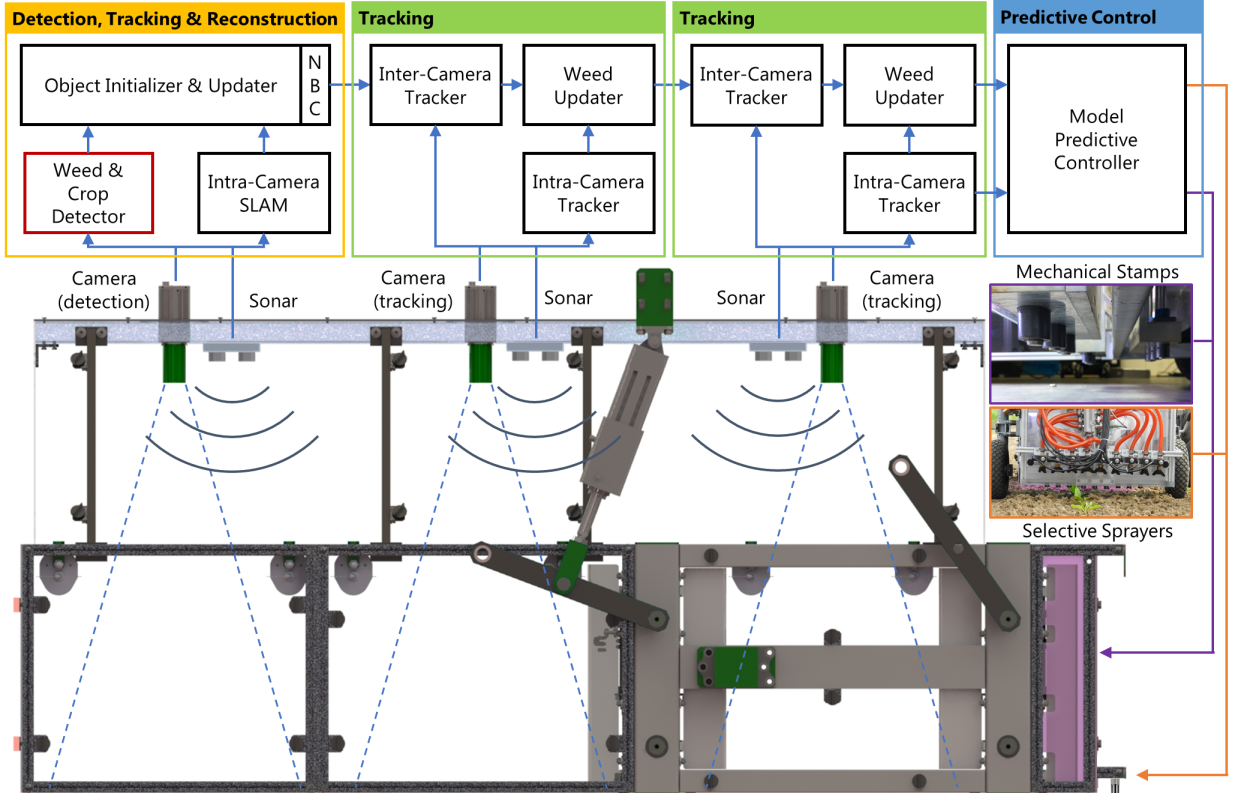}
   \caption{
         Bottom: Schematic 3D model of the weed intervention module 
      Top: An overview of our proposed weed control system that is composed of weed detection, tracking and predictive control modules. The weeds are tracked across the cameras and finally fed into a predictive control module to estimate the time and position of treatment at which they will be approaching the weeding tools.
   }
   \label{fig:bonirob-weed-unit}
\end{figure}

Supporting the target use case of selective weed intervention, the robot is equipped with an extension module, the Weed Intervention Module (\figref{fig:bonirob-weed-unit}). This module consists of a perception system for weed classification, multi-modal actuation systems, and their supporting aggregates.

The main design objectives of this unit are high weed throughput, precise treatment, and flexibility. 
The weeds are treated mechanically with two ranks of stampers or chemically with one rank of sprayers. The weeds are detected and tracked in real-time with three cameras with non-overlapping Field of View (FoV).

The perception system of the weed unit consists of three ground-facing global shutter cameras and three narrow beam sonars. To protect this perception system from natural light sources, the weed control unit is covered, and artificial lights have been installed to control the illumination. A first RGB+NIR camera is used for weed detection and tracking, while the other two RGB cameras are used for tracking. The sonars help recover the absolute scale of the camera images. Further details about the weed intervention module can be found in \secref{sec:sel_weeding}.

\section{Data Analysis and Interpretation in a Farming Scenario}\label{sec:data_analysis}

Precision farming applications aim to improve farm productivity while reducing the usage of fertilizers, herbicides, and pesticides. To meet these challenges, 
in-field measurements of plant vitality indicators and weed density are required. We addressed both these requirements from a robotic point-of-view, by proposing a set of methods to accurately detect plants and to distinguish them as crops and weeds (\secref{sec:_crop_weed_detection} and \ref{sec:synth_datasets}) and to automatically analyze the nitrogen status of crops from the multi-spectral aerial images (\secref{sec:n_status_detection}).

\subsection{Crop and Weed Detection}\label{sec:_crop_weed_detection}

\begin{figure}
   \centering
   \includegraphics[width=0.32\linewidth]{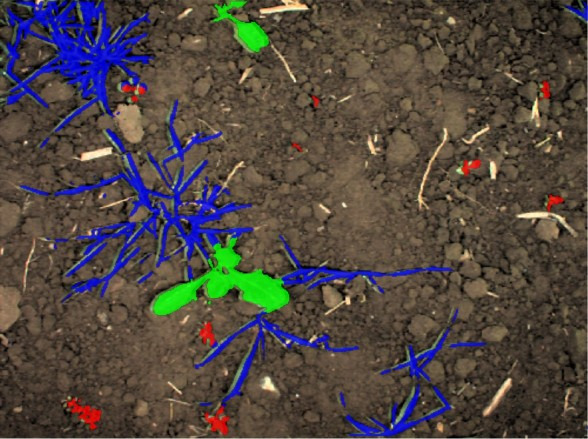}
   \includegraphics[width=0.32\linewidth]{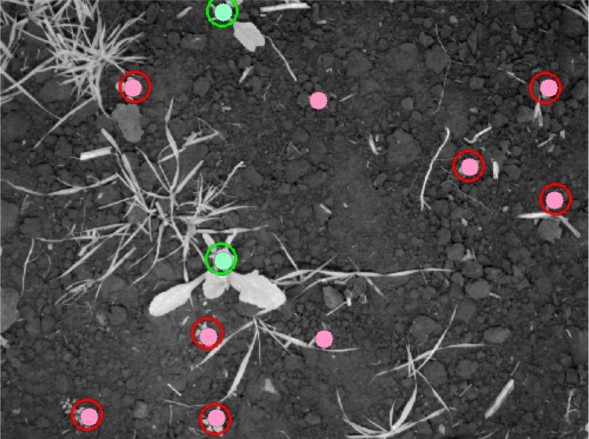}
   \includegraphics[width=0.32\linewidth]{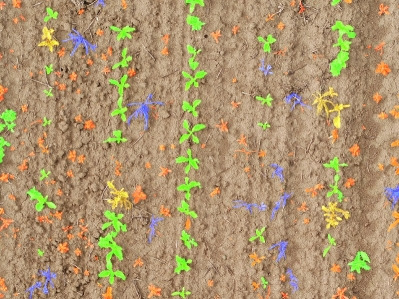}
   \caption{Example results obtained by our plant classification systems. Left: UGV-based semantic segmentation into crop, weed, grass-weed. Middle: stem detection providing accurate location of crops and weeds. Right: UAV-based semantic segmentation.}
   \label{fig:cwc}
\end{figure}

A prerequisite for selective and plant-specific treatments with farming robots is an effective plant classification system providing the robot with information on where and when to trigger its actuators to perform the desired action in real-time. 

In the Flourish project, we focus on vision-based approaches for plant classification and use machine learning techniques to effectively cope with the large variety of different crops and weeds as well as with changing environmental conditions.
\figref{fig:cwc} illustrates results obtained by our plant classification systems for both UGV and UAV platforms. The further distinction between weeds and grass-weeds allows our system to perform different treatments in a targeted manner depending on the type of weed. For example, local mechanical treatments are most effective when applied to the stem location of the plants. In contrast, grass-like weeds can effectively be treated by spraying herbicides to their leaf surfaces.

We developed several data-driven plant classification systems, ranging from approaches based on handcrafted features and Random Forests~\cite{lottes2017jfr,lottes2017icra} to deep learning approaches based on lightweight Fully Convolutional Networks (FCN)~\cite{lottes2018ral}. The latter showed superior performance and better generalization capability.

\begin{figure}
 \centering
 \includegraphics[width=1.0\linewidth]{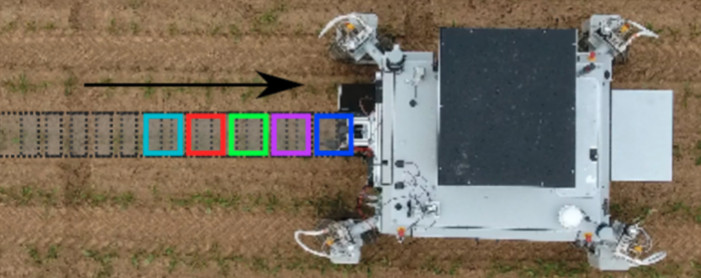}
 \includegraphics[width=1.0\linewidth]{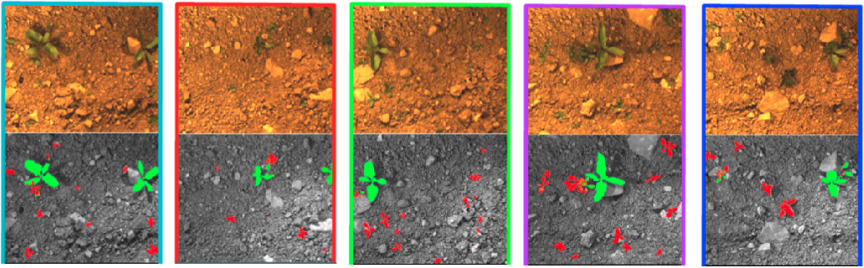}
 \caption{Top: The BoniRob UGV acquiring images while driving along the crop row.  Our approach~\cite{lottes2018ral} exploits an image sequence by selecting those images from the  history that do not overlap in object space; Bottom: Exemplary prediction of crop  plants and weed for the entire image sequence.  Note that the model was trained on data acquired in a different field.}
\label{fig:sequence}
\end{figure}

To effectively generalize to new conditions (different field, weather, \dots), we exploit geometric patterns that result from the fact that several crops are sown in rows. Within a field of row crops, the plants share a similar lattice distance along the row, whereas weeds appear randomly. In contrast to the visual cues, this geometric signal is much less affected by changes in visual appearance. We propose a semi-supervised online approach~\cite{lottes2017iros} by exploiting additional arrangement information of the crops to adapt the visual classifier. We also successfully tested approaches that operate on image sequences obtained along the crop rows, allowing the classifier to learn features describing the plant arrangement (\cite{lottes2018ral},~\figref{fig:sequence}).
The image sequence reveals that crops grow along the row and have a similar spacing, whereas the weeds appear randomly in the field strip. We show that incorporating this geometric information boosts the classification performance and the generalization capabilities of the plant classifiers.


The same underlying lightweight FCN structure is deployed on our UAV systems~\cite{sa2018ral}.
Here, we use the FCN in a classical single image fashion as the larger footprint of the camera implicitly covers enough information about the plant arrangement. Through our crop and weed classification systems, we enable UGVs to perform plant-specific high precision in-field treatments, and transform UAVs into an efficient system for crop monitoring applications. 

\subsection{Automatic Synthetic Dataset Generation}\label{sec:synth_datasets}

\begin{figure}[ht]
   \centering
   \includegraphics[width=\linewidth]{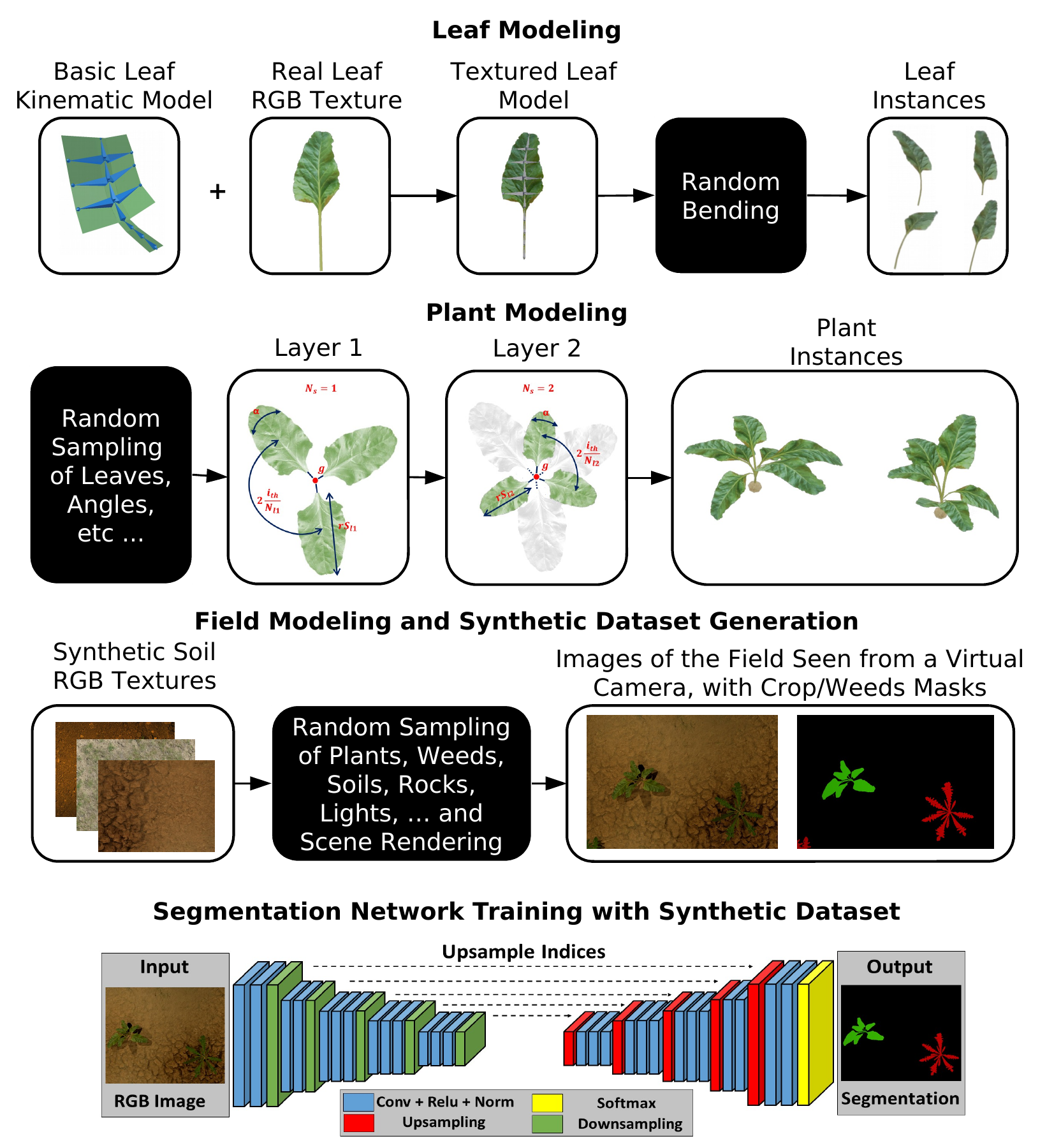}
   \caption{An overview of the automatic model based dataset generation procedure.}
   \label{fig:synth_dataset}
\end{figure}

High-performing data driven plant classification approaches usually require large annotated datasets, acquired across different plant growth stages and weather conditions. Annotating such datasets at a pixel-level is an extremely time consuming task.

We face this problem by proposing an automatic, model based dataset generation procedure \cite{dpgp_IROS2017} that generates large synthetic training datasets by rendering a large amount of photo-realistic views of an artificial agricultural scenario with a modern 3D graphic engine. To do so, we randomize a few key parameters (e.g., size, deformation and distribution of leaves, structure of plants, type of soil, \dots) with a bottom-up procedure (see \figref{fig:synth_dataset}):
\begin{itemize}
\item We model the leaves of the target plants using kinematic chains on which we apply a few real-world RGB textures;
\item We model the plants through radially distributed leaves layers, where the number of leaves per layer depends on plant species and the growth stage;
\item A virtually infinite number of realistic agricultural scenes can be then rendered by adding random soil backgrounds and by sampling random illumination conditions and plants distributions; the ground truth segmentation masks are automatically generated by the graphic engine;
\item The generated synthetic datasets can be used to effectively train modern deep learning based image segmentation architectures.
\end{itemize}
\subsection{Multi-spectral N-Status Detection and Phenotyping}\label{sec:n_status_detection}

The nutrition 
status of a crop is linked to yield formation and to 
the environmental footprint of agronomy. Well fertilized crops produce optimal yield and quality and are more stress resilient. Fertilizer deficiency hampers yield, whereas surplus supply increases the risk of nutrient loss to the environment and increases susceptibility to pests and diseases. Nitrogen plays a prominent role in the management of most crops, because of the generally high crop N demand and the very high mobility in the soil. Although sugar beet N demand is relatively low, yield and quality is strongly dependent on N- management: too low N-application limits tuber yield, while high N-application reduces the extractable sugar content in the tuber~\cite{Hofmann_2005}.

Therefore, it is important to apply N-fertilizer at the right time, rate and place. These decisions can be supported by optical remotes sensing tools making use of visible or non-visible parts of the spectral reflection of crop 
as used for N-status detection in Flourish project~\cite{Walter_ISPA_2018}. It serves as an example among others for the image-based assessment of plant traits, which play an increasingly prominent role for the development of sustainable agronomic practices in precision farming ~\cite{Walter6148, Finger_2019}. 

To validate the spectral and imaging methodology for sugar beets, in the Flourish project randomized field trials were established in commercial sugar beet fields and different nitrogen input treatments were applied from 2015 to 2017 . Aerial image spectroscopy was realized with a multi-spectral Gamaya VNIR 40 camera mounted on a UAV. As ground truth plant N status, tuber yield and sugar content were measured. As for other crops, our results show red edge based spectral indices such as the simple ratio and the normalized difference red edge ratio~\cite{Argento_2019_ECPA} indicating the N-status in sugar beets successfully from the UAV-based sensor, resulting in useful N-fertilizer application maps.

\section{Positioning and Environment Modeling}\label{sec:env_modeling}

The ability to localize and build a model of the surrounding environment is an essential requirement to support reliable navigation of an autonomous robot. Such tasks are even more challenging in a farming scenario, where (a) the environment is mainly composed of repetitive patterns, with no distinctive landmarks; (b) multi-spectral information should be included in the modeling process, to support decision making for farm management. Moreover, in a multi-robot setup as in the Flourish project, the UAV and the UGV should be able to  cooperatively build a shared model of the environment. This section presents the main contributions we proposed to localize and model cultivated fields by a UAV (\secref{sec:uav_localization}), a UGV (\secref{sec:ugv_localization}), and to fuse this information between robots (\secref{sec:cooperative_modeling}) and across time (\secref{sec:temporal_registration}).

\subsection{UAV Localization and Mapping}\label{sec:uav_localization}



The aim of the UAV perception system is to collect high-resolution spatio-temporal multi-spectral maps of the field. This data is critical as it allows for mission planning before the UGV actually enters the field, thereby optimizing the time/location of ground intervention procedures without the risk of crop damage and soil compaction. The perception pipeline requires two main competencies: (1) motion estimation and precise localization within the field and (2) multi-resolution multi-spectral aerial mapping based on the indicators needed to assess plant health.

\begin{figure}
  \centering
  \begin{minipage}[b]{\linewidth}
      \includegraphics[width=\columnwidth]{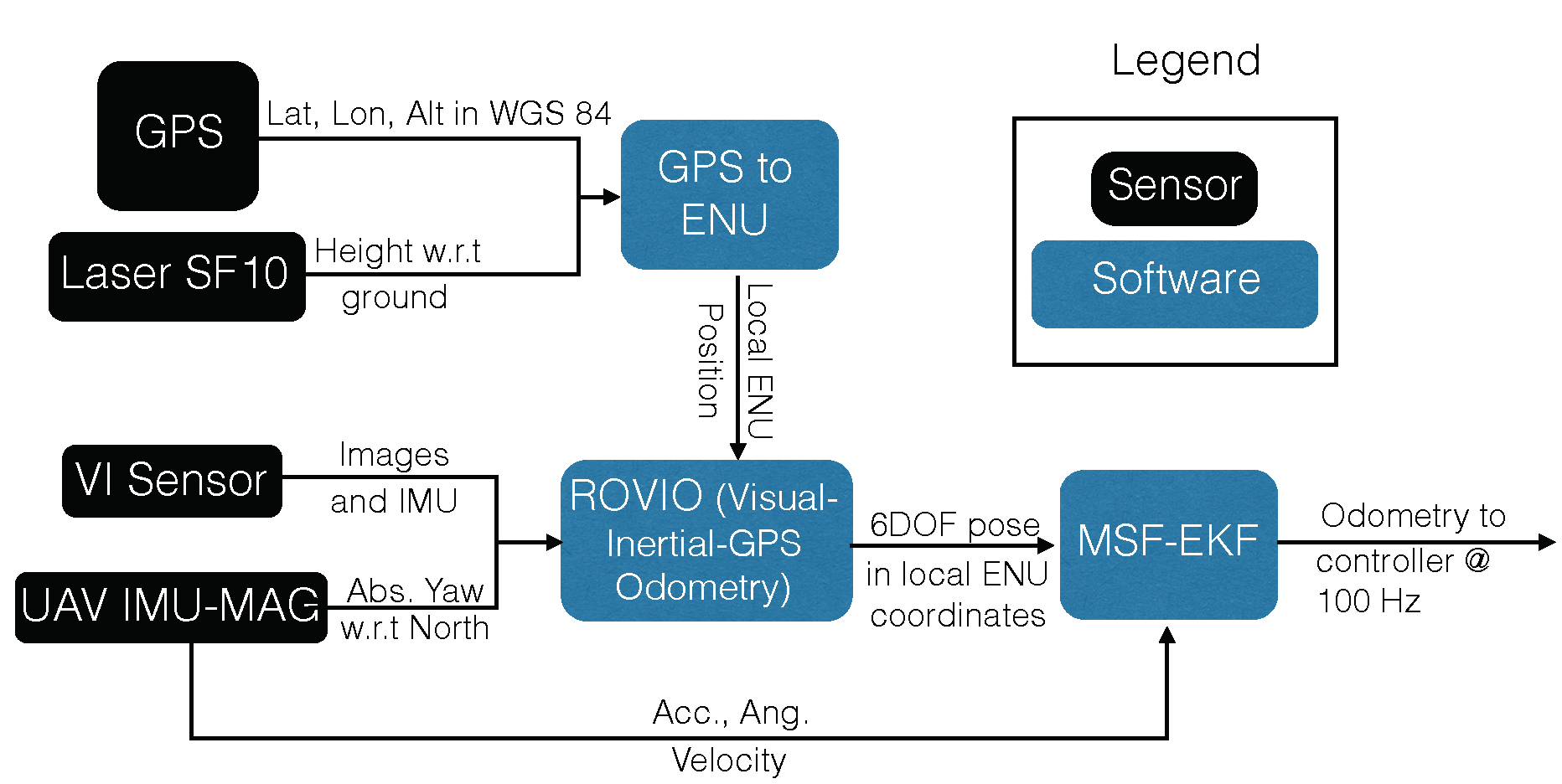}
  \end{minipage}
  \begin{minipage}[b]{\linewidth}
    \includegraphics[width=\columnwidth]{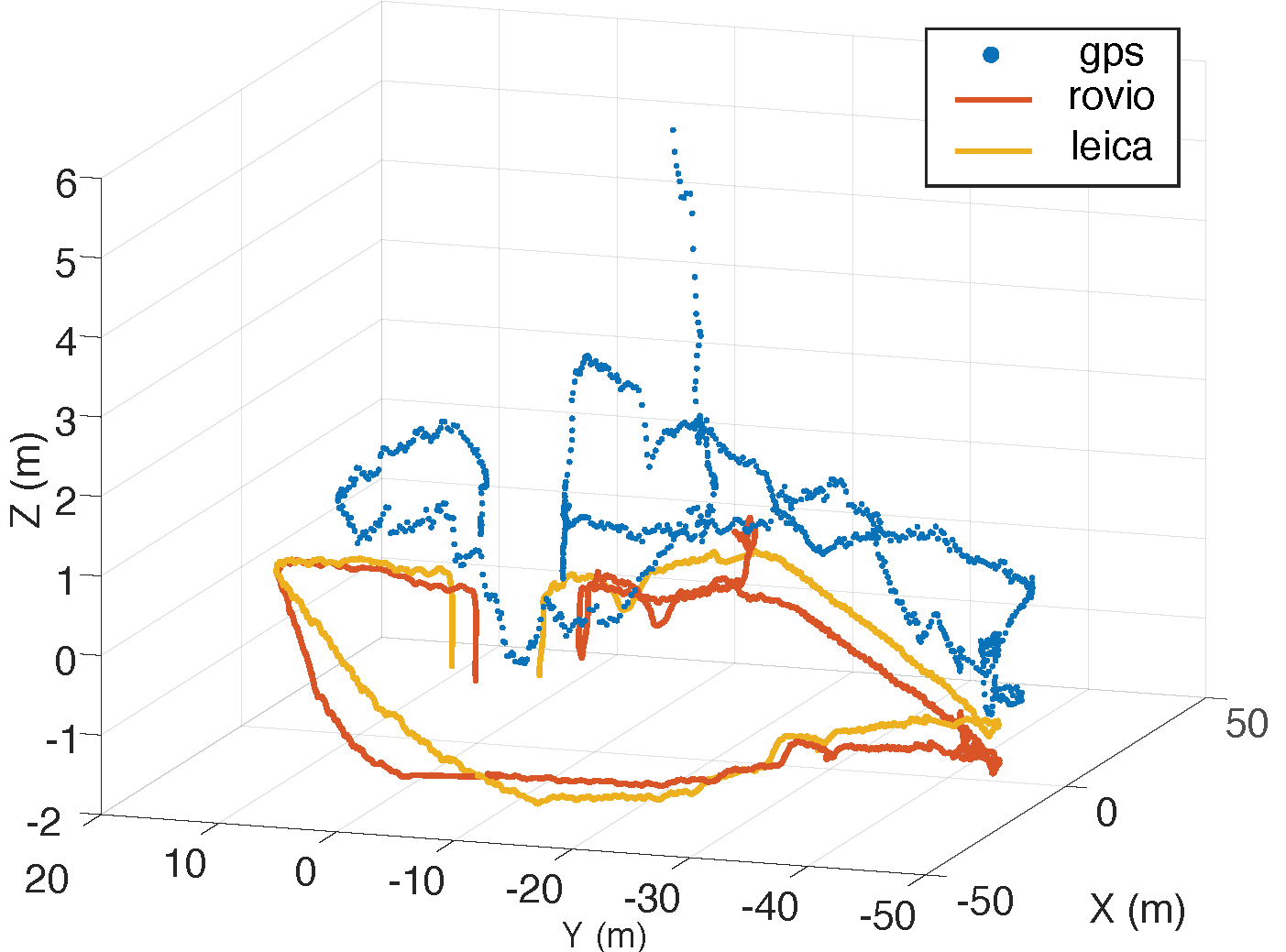}
  \end{minipage}
      \caption{Top: Block diagram of our UAV state estimation framework depicting the sensor suite and major software components. Bottom: Comparison of our VI-GPS fusion-based state estimation with raw GPS and ground truth.}
      \label{fig:uav_localization}
\end{figure}

Key challenges for on-field vision-based localization are the homogeneous appearance of crops and the accuracy of the GPS which, used alone, is not sufficient to construct maps for defining paths for UGV intervention. To address this, we develop an on-board state estimation system that combines data from a synchronized VI sensor, a GPS sensor, the UAV IMU and, optionally, a laser altimeter, to estimate the 6 DoF pose. \figref{fig:uav_localization} (top) overviews the system. The Robust Visual Inertial Odometry (ROVIO)~\cite{Bloesch2015} framework is used to produce a 6 DoF pose output based on raw images and IMU data from the VI sensor. The Multi-Sensor Fusion (MSF) framework~\cite{Lynen2013} then combines the ROVIO output and the UAV IMU data to obtain state estimates passed to our Model Predictive Controller (MPC) for trajectory tracking. To improve accuracy and robustness, we integrate our system with MAPLAB~\cite{Schneider2018}, a framework with map maintenance and processing capabilities. On-field results using an AscTec NEO and DJI Matrice 100 UAV platforms demonstrate high state estimation accuracy compared to ground truth from a Leica Geosystems Total Station (\figref{fig:uav_localization}, bottom).

\begin{figure}
  \centering
  \includegraphics[width=\columnwidth]{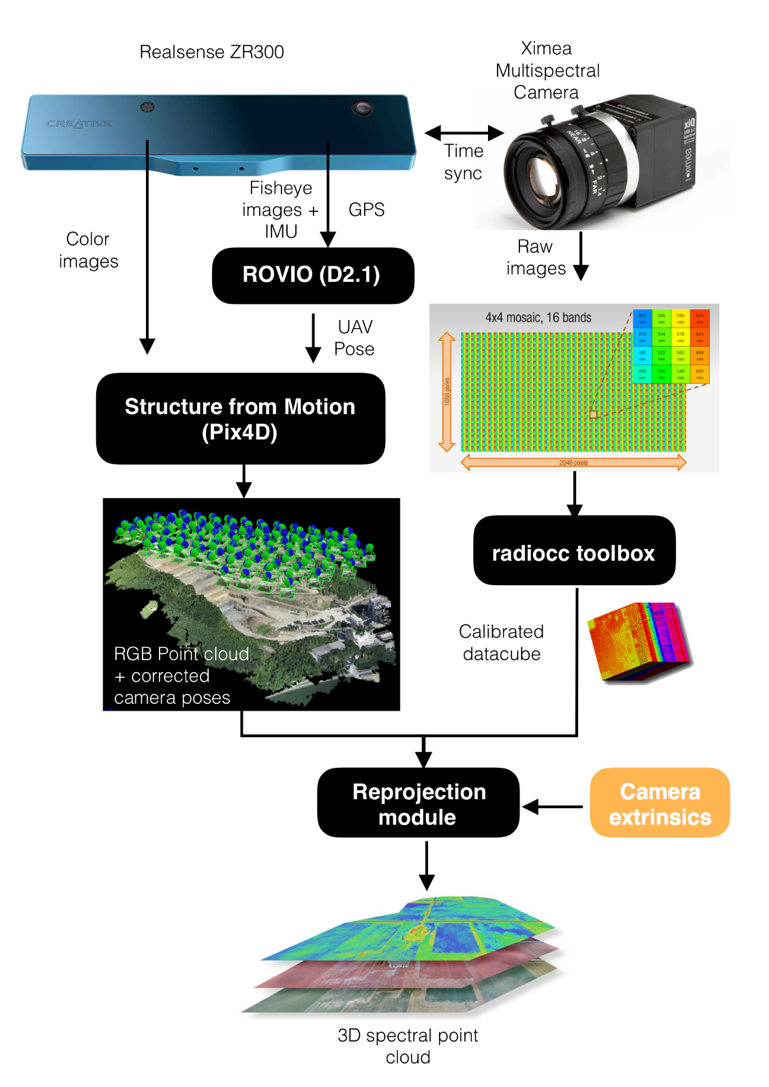}
  \caption{Block diagram of our environment modeling framework depicting the sensor suite and major software components (images courtesy of Intel\textregistered, Pix4D S.A and Ximea).}
\label{fig:env-model-pipeline}
\end{figure}

\begin{figure*}
  \centering
  \begin{minipage}[b]{0.49\linewidth}
    \includegraphics[width=\linewidth]{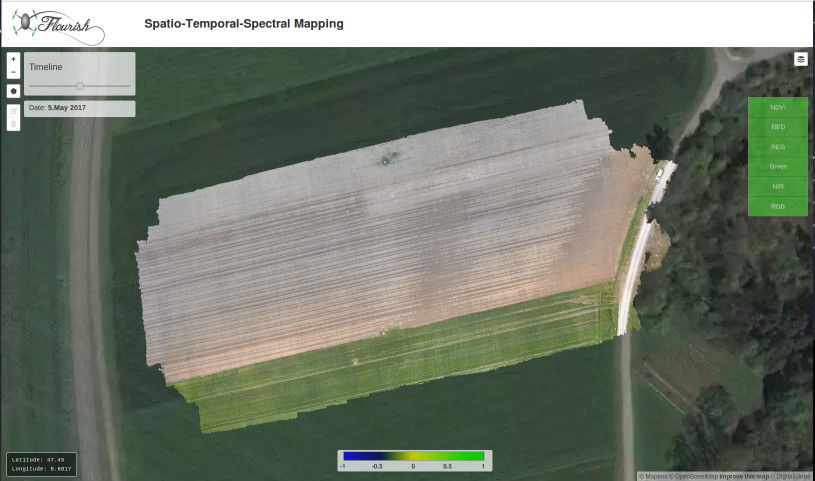}
  \end{minipage}
  \begin{minipage}[b]{0.49\linewidth}
  	\includegraphics[width=1\columnwidth]{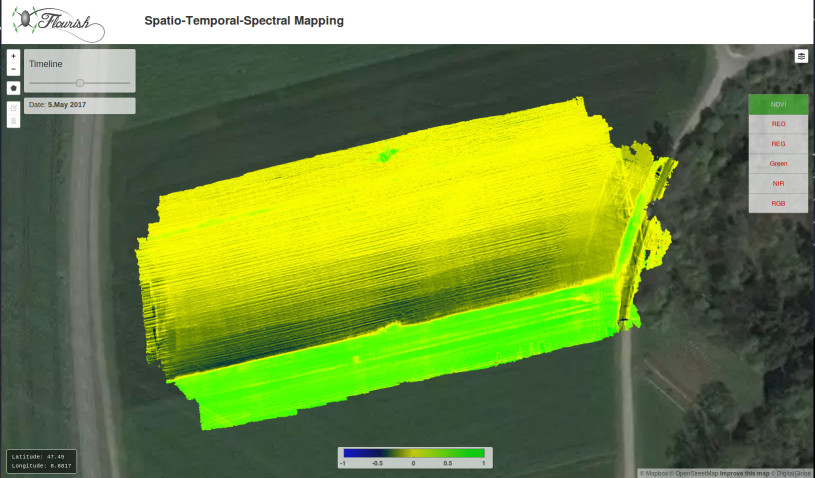}
  \end{minipage}
  \caption{Visualization interface for the spatio-temporal-spectral database showing both RGB orthomosaics (left) and corresponding index maps (right) for a sugar beet field over time. The user can select spectral layers to view a georeferenced reflectance orthomosaic corresponding to a wavelength band, view the color orthomosaic, and toggle through all available surveys using the timeline.}
\label{fig:visualization-interface}
\end{figure*}

High-resolution field map models are a key prerequisite for enabling robots in precision agriculture. To this end, we develop a UAV environmental modeling framework using the pose estimate from our localization system, and color and multi-spectral camera information over multiple flights, to create spatio-temporal-spectral field models. \figref{fig:env-model-pipeline} shows our pipeline. Taking raw RGB and multi-spectral images, and UAV poses as inputs, we radiometrically correct spectral data to create a spatial field model in the form of a dense point cloud. For each point, the spectral reflectances in the multi-spectral wavelength bands are estimated and stored. The field evolution over time can be viewed through layered orthomosaics generated from this data through a custom browser-based visualization module (\figref{fig:visualization-interface}). We use higher-quality RGB camera images, recovering high-resolution field geometry, and use the relative position and orientation between the RGB and multi-spectral camera to estimate its spectral reflectance. Importantly, our strategy removes the need for a separate reconstruction step for each band and the subsequent alignment step.

\subsection{UGV Global Positioning and Mapping}\label{sec:ugv_localization}

\begin{figure}[t!]
   \centering
   \includegraphics[width=\linewidth]{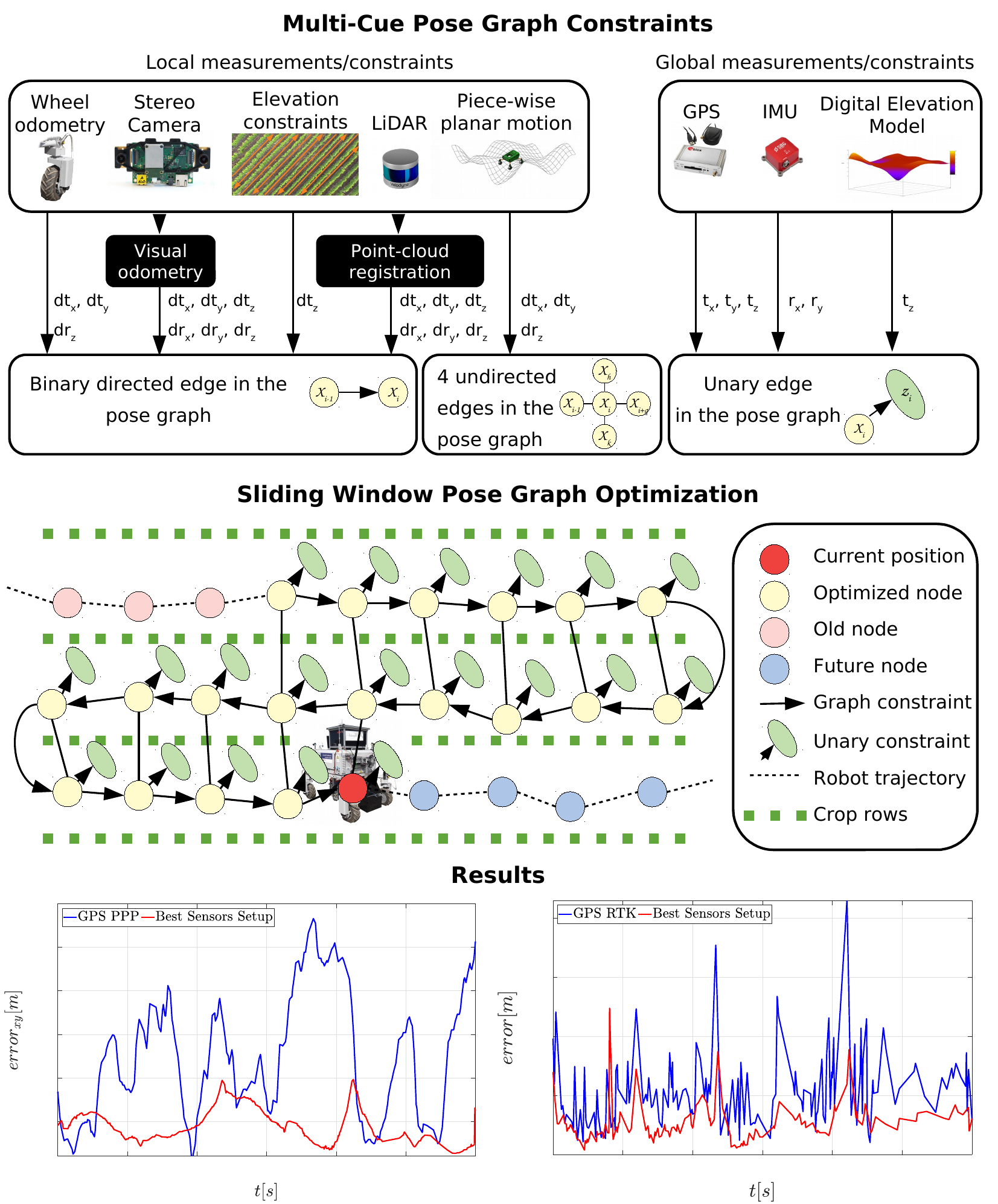}
   \caption{Overview of the proposed graph based multi-cue UGV positioning system. Top: the constraints (i.e., the sensors and provided measurements) integrated in the pose graph; Middle: the conditional dependencies between nodes; Bottom: absolute error plots for the raw GPS trajectory (blue) and the optimized trajectory (red), for two types or GPSs.}
   \label{fig:positioning}
\end{figure}

Currently, most positioning systems used in commercial farming UGV rely on high-end Real-Time Kinematic Global Positioning Systems (RTK-GPSs) that, however, are not robust enough against base station signal loss or multi-path interference and cannot provide the full 6D position (translation and rotation) of the vehicle. We tackle this problem by proposing a UGV positioning system \cite{ipngp_RA-L2018} that effectively fuses several heterogeneous cues extracted from consumer-grade sensors and exploits the specific characteristics of the agricultural context with a few additional constraints. We formulate the global localization problem as a 6D pose graph optimization problem. The constraints between consecutive nodes  (\figref{fig:positioning}, top) are represented by motion estimations (wheel odometry, visual odometry, \dots). Noisy, but drift-free GPS and IMU readings are directly integrated as prior nodes. Driven by the fact that both GPS and VO provide poor estimates along the $z$-axis (i.e., parallel to the gravity vector), we introduce two additional altitude constraints:
\begin{enumerate}
 \item An altitude prior, provided by a Digital Elevation Model (DEM);
 \item A smoothness constraint for the altitude of adjacent nodes.
\end{enumerate}

The integration of such constraints improves the accuracy of the altitude estimation and also benefits the estimation of the remaining state components. The optimization problem is cyclically solved online by using a sliding window strategy (\figref{fig:positioning}, middle).

\subsection{Cooperative UAV-UGV Environment Modeling}\label{sec:cooperative_modeling}

\begin{figure*}[ht!]
   \centering
   \includegraphics[width=\linewidth]{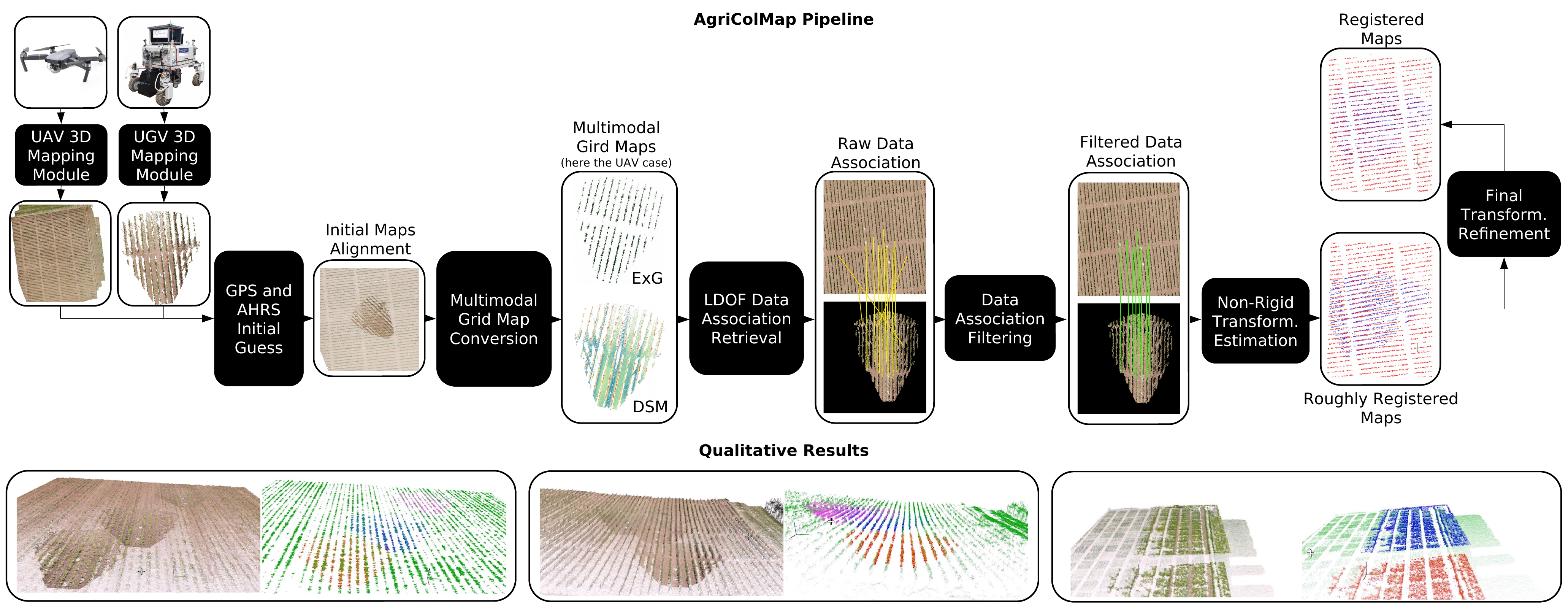}
   \caption{Top: An overview of the AgriColMap method pipeline; Bottom: Some qualitative registration results (RGB and $ExG$-filtered point clouds). The UGV clouds are clearly visible due to their higher points density.}
   \label{fig:agricolmap}
\end{figure*}

Building a shared map of the environment is an essential but challenging task: the UAV can provide a coarse reconstruction of large areas, that should be updated with more detailed information collected by the UGV. We introduced AgriColMap~\cite{pknsnp_RA-L2019}, an acronym for Aerial-Ground Collaborative 3D Mapping for Precision Farming, an effective map registration pipeline that registers heterogeneous maps built by the UGVs and UAVs.

AgriColMap leverages a multimodal field representation and formulates the data association problem as a large displacement dense optical flow (LDOF) estimation. The complete pipeline is schematized in \figref{fig:agricolmap}. We assume that both the UAV and the UGV can generate colored, georeferenced point clouds of a farm environment, $\mathcal{M}_A$ and $\mathcal{M}_G$, e.g., using photogrammetry-based 3D reconstruction. Our goal is to estimate an \emph{affine} transformation $F:\mathbb{R}^3 \rightarrow \mathbb{R}^3$ that allows to accurately align them by compensating the geo-tags misalignments and the reconstruction and scale errors. We start looking for a set of point correspondences, $m_{A,G} = \{(p, q) : p \in \mathcal{M}_A, q \in \mathcal{M}_G\}$, that represent points pairs belonging to the same global 3D position. Inspired by the fact that points in $\mathcal{M}_A$ locally share a coherent ''flow`` towards corresponding points in $\mathcal{M}_G$, we cast the data association problem as a \textit{dense}, \textit{regularized}, matching approach. This problem recalls the dense optical flow estimation problem for RGB images: we introduce a multimodal environment representation that allows to exploit such 2D methods on 3D data, while enhancing both the semantic and geometrical properties of the maps. We exploit two intuitions:
\begin{itemize}
\item A Digital Surface Model (DSM)
well highlights the geometrical properties of a cultivated field;
\item A vegetation index can highlight the meaningful parts of the field and its visual patterns.

\end{itemize}
We transform $\mathcal{M}_A$ and $\mathcal{M}_G$ into 2D grid maps $\mathcal{J}_A,\mathcal{J}_G:\mathbb{R}^2 \rightarrow \mathbb{R}^2$, where for each cell $p$ we provide the surface height $h$ and the Excess Green index, $ExG(p) = 2g_p - r_p - b_p$, being $r_p,~g_p,~b_p$ the (average) RGB components of the cell.
To estimate the offsets map, we employ a modified version of the LDOF Coarse-to-fine PatchMatch framework (CPM)~\cite{hu2016}. We apply the visual descriptor of the original CPM method directly to the  ExG channel of $\mathcal{J}_A$ and $\mathcal{J}_G$, while we exploit a \emph{3D descriptor} computed over the DSM to extract salient geometric information; the matching cost has been modified accordingly to take into account both descriptors.

The largest set of coherent flows defines a set of matches $m_{A,G}$ that are used to infer a preliminary alignment $\hat{F}$. We finally estimate the target affine transformation $F$ by exploiting the Coherent Point Drift registration algorithm~\cite{myronenko2010}, over the point clouds $\mathcal{M}^{veg}_A$ and $\mathcal{M}^{veg}_G$, that are obtained from $\mathcal{M}_A$ and $\mathcal{M}_G$  by extracting only points that belong to vegetation with an ExG based thresholding operator.

\subsection{Long-Term Temporal Map Registration}\label{sec:temporal_registration}

\begin{figure*}[ht!]
     \centering
      \includegraphics[width=\linewidth]{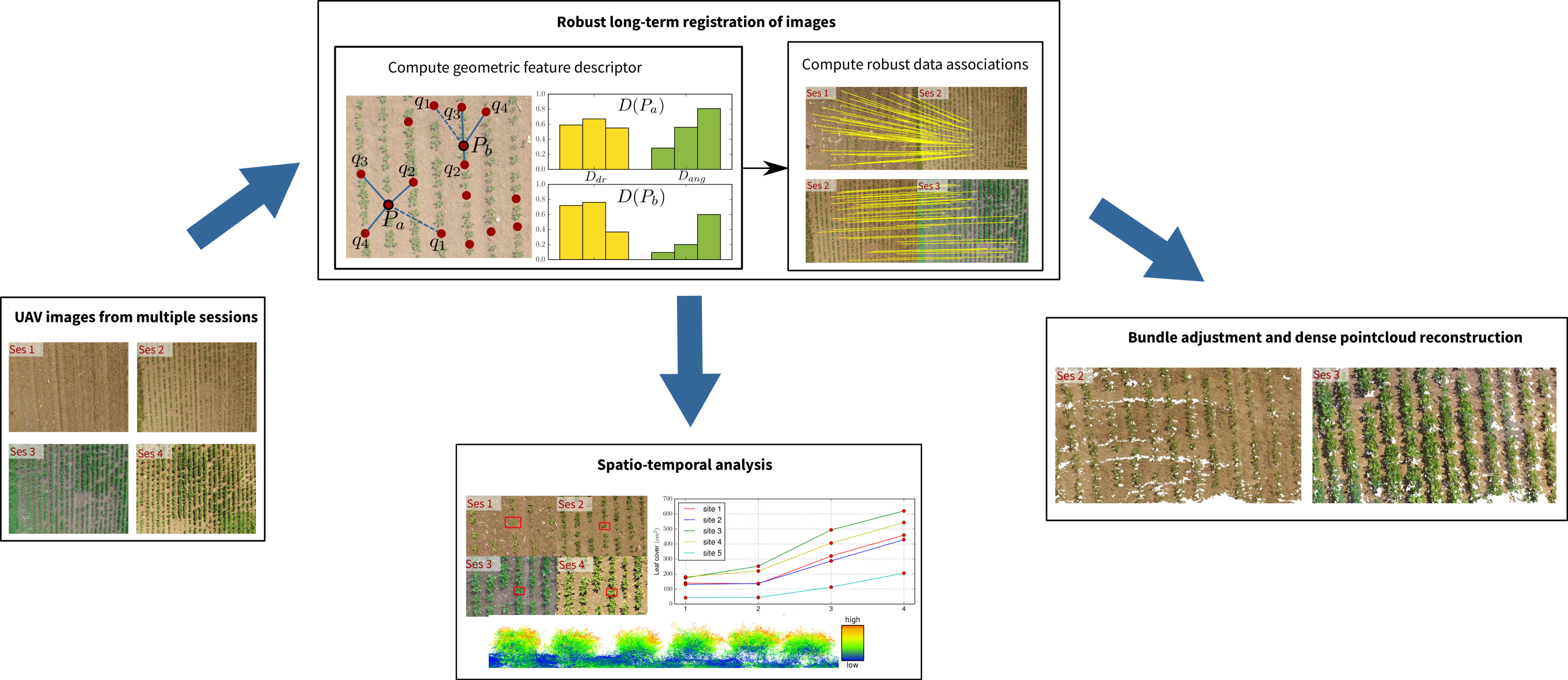}
      \caption{Pipeline for performing long-term registration of UAV images of crop fields. The registration approach computes robust data associations between temporally separated images based on a scale-invariant geometric descriptor. These associations allow for registering the data in a common reference frame, which then forms the basis for performing spatio-temporal analysis.}
    \label{fig:long_term_registration_pipeline}
\end{figure*}

Continuous crop monitoring is an important aspect of phenotyping and requires the registration of sensor data over the entire season. This task is  challenging due to the strong changes in the visual appearance of the growing crops and the field itself. Conventional image registration based on visual descriptors is typically unable to deal with such drastic changes in appearance. To address this challenge, we developed a method for registering temporally separated images by exploiting the inherent geometry of the crop arrangement in the field, which remains relatively invariant over the season. We propose a scale-invariant, geometric feature descriptor that encodes the local plant arrangement geometry and uses these descriptors to register the images even in the presence of strong visual changes~\cite{chebrolu2018ral}.
The registration results allow for spatio-temporal analysis of data collected over the crop season. This includes applications such as monitoring growth parameters at a per plant level, as illustrated in~\figref{fig:long_term_registration_pipeline}.

\section{Planning, Navigation and Coordination}\label{sec:planning_navigation}

The UAV and the UGV have different working areas and roles within each field analysis and targeted intervention mission. Their action planning and navigation policies should reflect these differences. We introduced an ad-hoc UAV navigation module (\secref{sec:uav_planning}) using a planner to effectively perform field monitoring missions while respecting battery constraints. Crop row localization and safe in-field UGV navigation is addressed in \secref{sec:ugv_navigation}, where the high number of DoFs of the UGV is used to improve motion efficiency and smoothness. The inter-robot mission coordination framework is then introduced in \secref{sec:mission_coordination}.

\subsection{UAV Mission Planning and Navigation}\label{sec:uav_planning}

A key challenge in agricultural monitoring is developing mission planning algorithms to define the path for a UAV to optimally survey the field. The planning module needs to maximize mapping accuracy given battery life constraints, taking into account field coverage and scientifically defined areas of interest. We developed an informative path planning (IPP) framework for adaptive mission planning to cater for these requirements~\cite{Popovic2020}.

Our framework is suitable for mapping 
a terrain depending on the type of data received from an onboard sensor, e.g., a depth or multi-spectral camera.
In terms of mapping, the main challenge is fusing the dense visual imagery into a compact probabilistic map in a computationally efficient way. To address this, we present a new method for multiresolution mapping that considers the patterns of the target distributions on the farm. We use Gaussian Processes (GPs) to encode the spatial correlations common in biomass distributions. A GP model is exploited as a prior for recursive Bayesian data fusion with probabilistic, variable-resolution sensors. In doing so, our approach enables mapping 
without the computational burden of standard GP regression, making it suitable for online, on-platform applications.

In terms of planning, a fundamental problem we tackle is trading off image resolution and FoV to find the most useful measurement sites at different flying altitudes. During a mission, the terrain maps built online are used to plan trajectories in continuous 3D space that maximize an information-based objective
, e.g., targeted high-resolution mapping of areas 
infested by weeds. Our planning scheme proceeds in a finite-horizon fashion, alternating between replanning and plan execution. This allows us to create \textit{adaptive} plans, taking new sensor data into account online to focus on areas of interest as they are discovered. For replanning, we leverage an evolutionary technique, the Covariance Matrix Adaptation Evolution Strategy (CMA-ES), to optimize initial trajectory solutions obtained by a course 3D grid search in the UAV workspace. 

\begin{figure}[!t]
\centering
  \includegraphics[width=\linewidth]{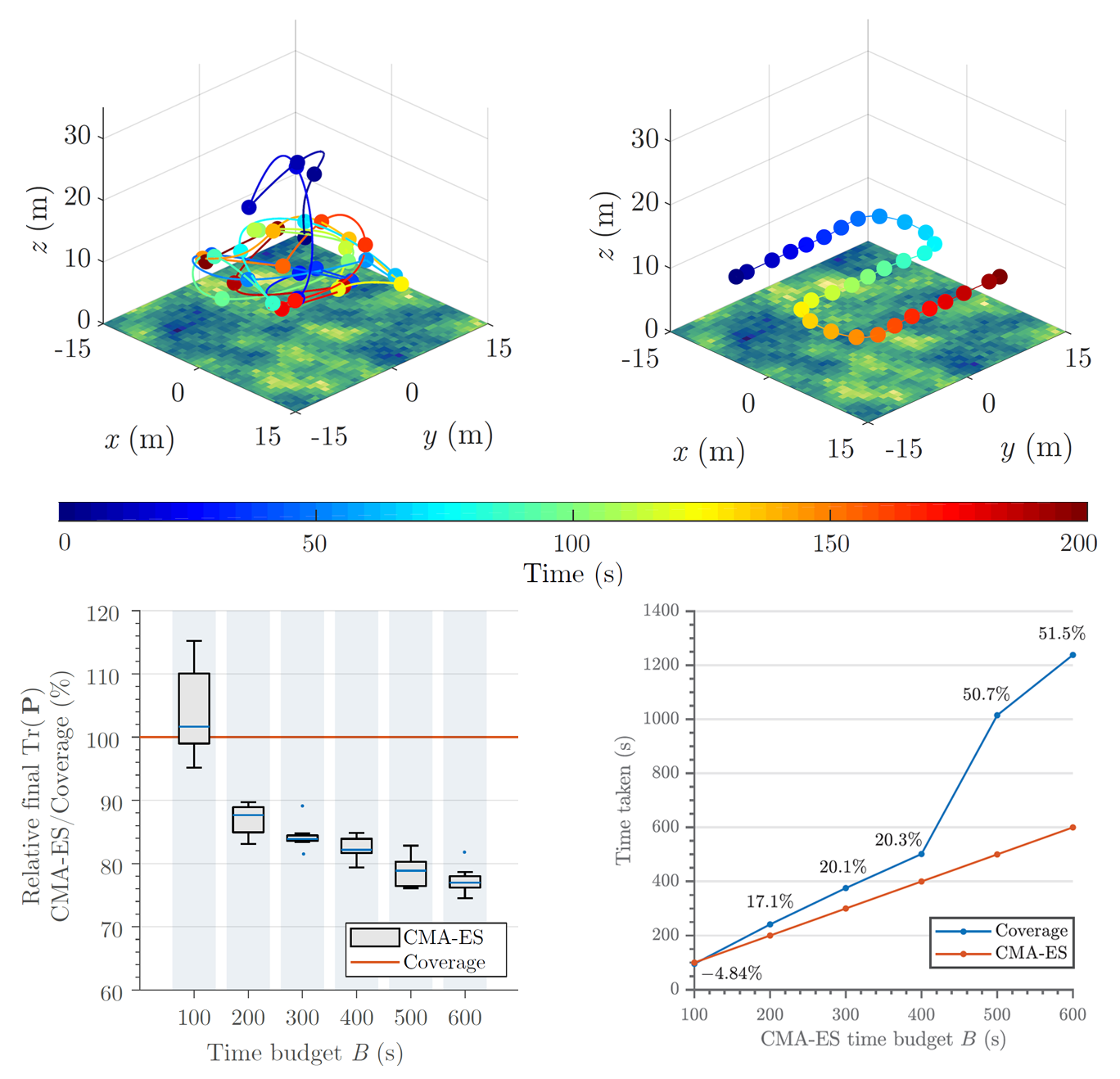}
\caption{
Top: Example comparison of our CMA-ES-based approach to ``lawnmower'' coverage (left and right, respectively) for 
mapping in $200$\,s missions. The colored lines and spheres represent the traveled trajectories and measurement sites. Ground truth maps are rendered. Bottom-left: Comparison of the final map uncertainties (measured by the GP covariance matrix trace) for various path budgets. Ten CMA-ES trials were run for each budget. Bottom-right: Comparison of times taken to achieve the same final map uncertainty, given a fixed CMA-ES budget. 
} \label{fig:ipp_results}
\end{figure}

Our approach was evaluated extensively in simulation, where it was shown to outperform existing methods (\figref{fig:ipp_results}), and validated on the field. 

\subsection{UGV Position Tracking and Navigation}\label{sec:ugv_navigation}
\begin{figure*}
  \centering
  \includegraphics[height=7cm]{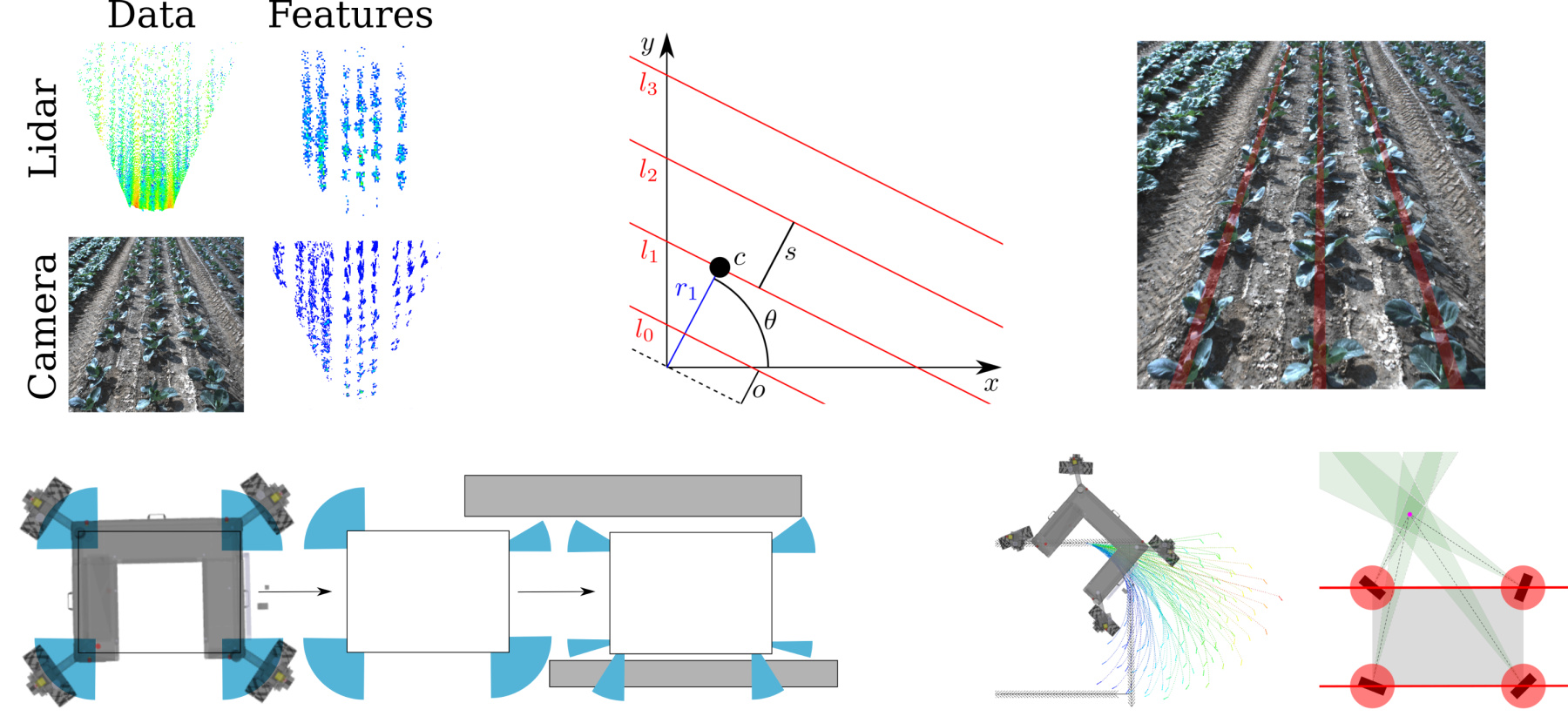}
    \caption{Overview of the UGV navigation system.\quad Top left: The
      Pattern Hough Transform detects crop rows in lidar or
      camera data using the extracted plant features.\quad Top center: A
      pattern is defined as set of parallel and equidistant lines
      (red) with orientation $\theta$, offset $o$ and spacing
      $s$.\quad Top right: Result of the Pattern Hough Transform (red) on
      pointed cabbage ($\approx 10$cm). \quad Bottom left:
      Valid arm angle intervals when moving close to an
      obstacle. \quad Bottom center: Velocity rollouts in the local
      planner.  The rollouts are color coded with their respective
      costs. Bottom right: The ICR constraints derived from the
      hardware constraints (red) and the maximum steering velocity
      (green).}
  \label{fig:navigation}
\end{figure*}

For autonomous navigation on fields the BoniRob UGV needs to accurately steer along the crop rows without crushing
any of the value crop. Moreover, to transition between crop
rows, performing tight and accurate turns at the end of the field is
essential.
There are three key requirements to achieve this: a pose
  estimate relative to the rows, a path along the crop rows through
  the field, and smooth velocity commands to precisely follow this
  path.
To this end, we developed a crop row detection algorithm, the
\emph{Pattern Hough Transform}~\cite{winterhalter18ral}. We first
process the input from vision or lidar data by extracting plant
features and projecting them onto a feature grid map in the local
robot frame (see \figref{fig:navigation}, top left).  Then, our
  Pattern Hough Transform determines the pattern of parallel and
  equidistant lines that is best supported by the feature map (see
  \figref{fig:navigation}, top right).
Such a pattern is defined by the orientation $\theta$, the
  spacing between adjacent lines $s$ and the offset of the first line
  to the origin $o$ as shown in \figref{fig:navigation} (top center).
Since our approach takes into account all available data to
  detect the crop rows in a single step, it is robust against outliers
  like weed growing between the crop rows, and yields accurate
  results during turning, i.e., when the robot is not necessarily
  aligned with the crop rows.

We integrated the output from our Pattern Hough Transform into the
localization module of the BoniRob. The localization is based on an
\emph{Extended Kalman Filter}. We use fused odometry and IMU
  measurements for the prediction.
In the correction step, we align the detected
crop row pattern with a GPS-referenced map of crop rows to correct the
pose estimate of the robot relative to the field. Since the crop row
pattern only provides lateral and orientation information, i.e., no
correction along the crop rows, we correct the
longitudinal position estimate with GPS signals.

We implemented a global planner based on a state lattice planner to
ensure that the BoniRob finds a path to any reachable pose in the
field.  The BoniRob can change its track width by adjusting
  the angles of the lever arms to which the wheels are attached
  (\figref{fig:uav_ugv}, right). Thus, whether it can pass through a
  narrow gap or over an obstacle depends on the wheel positions (see
  \figref{fig:navigation} bottom left).
We developed a path planner that considers the lever angles
explicitly~\cite{fleckenstein17icra} by including the arm
  angles in the state space and adding actions that allow the planner
  to change them. Adding the arm angles greatly increases the size of
  the state space, which makes planning with commonly used search
  algorithms inefficient.
Thus, we introduced a novel method to represent the robot state with a
reduced cardinality, that is, we track valid arm angle intervals
instead of single arm angles in the robot state.

Our local planner translates a pose path from the global
  planner into velocities while considering steering constraints.
Any robot with slow-turning independently steerable wheels, such as
the BoniRob, has certain steering constraints.  The most
  prominent constraints are the limited steering velocity,
  non-continuous steering or wheel angle instabilities when the center
  of rotation is on a wheel.
To avoid violations of these steering constraints, we
  presented a new approach that incorporates steering constraints when
  generating velocity rollouts (see \figref{fig:navigation} bottom
  right)~\cite{fleckenstein19fsr}.
Our approach leverages
the correspondence between the wheel angles and the
instantaneous center of rotation (ICR) of the robot. After projecting
the steering constraints into ICR space, we compute a valid ICR path
that satisfies
the constraints. From this ICR path, we calculate valid velocity
sequences that the robot can execute smoothly. Real-world experiments
show that our local planner improves efficiency and leads to smoother
execution.

\subsection{UAV-UGV Mission Coordination}\label{sec:mission_coordination}

To unlock the potential of the Flourish robotics system, it is essential to be able to run coordinated missions between the robots.
Since both robots share information via Wi-Fi, this information needs to be kept at a minimal level and the coordination needs to be ensured even when communication is lost. The only data exchanged are: the UAV pose, the UGV pose, the coordinates of the areas of interest, the requests from one robot to the other and their status messages. Because of the lossy communication, exchanging requests and status is a reliable way to ensure a message sent by a robot is indeed received by the other one.

\begin{figure}[htb]
   \centering
  \includegraphics[width=\linewidth]{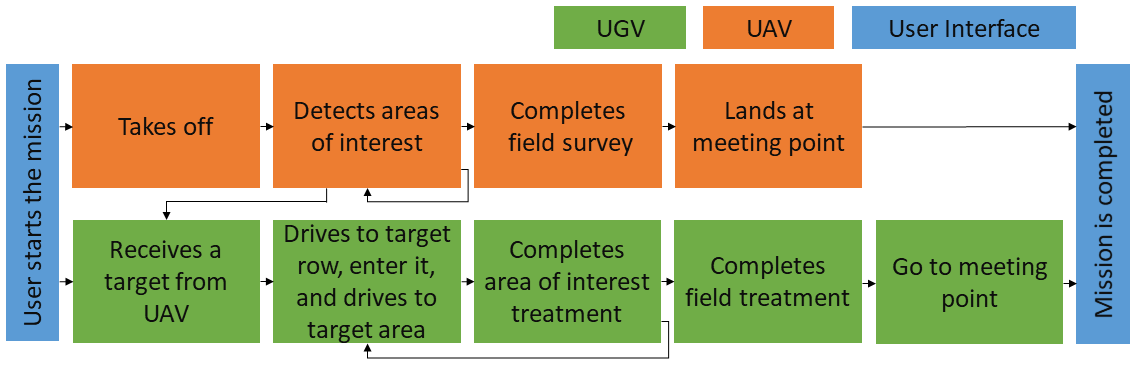}
   \caption{
      Example of a coordinated mission.
  }
   \label{fig:coordination}
\end{figure}

The mission framework used on both robots is based on ros\_task\_manager~\cite{pradalier:hal-01435823}. This is a task scheduler developed for ROS particularly easy to use, that allows for combining multiple behaviors, with elements running in sequence or in parallel, eventually interrupting each other. This framework is based on tasks implemented in C++, which are combined into complex missions implemented in basic Python.
\figref{fig:coordination} illustrates an example of coordinated mission.

\section{In-Field Intervention: the Collaborative  Weeding  Use Case}\label{in-field_intervention}

The main use case addressed in the Flourish project is the \emph{collaborative weeding application} (\figref{fig:concept}). The UAV flies over the field running the navigation and planning algorithms of \secref{sec:uav_localization} and \ref{sec:uav_planning}, while analyzing the weed pressure by using the classification algorithms presented in \secref{sec:_crop_weed_detection}. High weed pressure areas are notified to the UGV by using the coordination framework described in \secref{sec:mission_coordination}. Thus, the UGV starts to move toward the selected areas, running the algorithms of \secref{sec:ugv_localization} and \ref{sec:ugv_navigation}. In this section, we describe the tools (\secref{sec:sel_weeding}) and methods (\secref{sec:weed_tracking}) used for the actual weed treatment, with possible agronomic impacts reported in \secref{sec:agronomic_impact}.
We successfully tested the whole pipeline in a public demonstration during a dissemination event held near Ancona (Italy) in May 2018.

\subsection{Selective Weed Removal}
\label{sec:sel_weeding}

The weed  intervention  module (\figref{fig:bonirob-weed-unit}), whose perception system was introduced in Sec.~\ref{sec:bonirob}, includes further tools designed to address the targeted weed treatment: a weed stamping tool and a selective spraying tool (\figref{fig:bonirob-weed-unit}). The stamping tool is composed of 18 pneumatic stamps arranged in two ranks. All stamps are individually controllable and a high precision of the positioning is ensured by only allowing one degree of freedom for the positioning across to the driving direction. The spraying tool is positioned in the back. It is assembled out of nine nozzles, individually controlled by off the shelf magnetic valves. 

Both weeding tools are controlled with a scalable, programmable logic controller (PLC). Modules requiring more computational resources, i.e weed detection and tracking, are implemented on a computer dedicated to the weed control running Linux and ROS. 

The bolt of the stamps have a 10mm diameter, whereas the footprint of a sprayer is 30mm when set in the lowest position as in our experiments.  To actually treat a weed while the robot is moving is a time-critical part of the process because a small delay can lead to a position error at centimeter-level that is large enough to miss a small weed. 
In our experiments, the decision on which tool is used on which weed is only based on a size criteria: large weeds are sprayed while small weeds are stamped.

\subsection{Weed Tracking}\label{sec:weed_tracking}

\begin{figure}[t]
   \includegraphics[width=\linewidth]{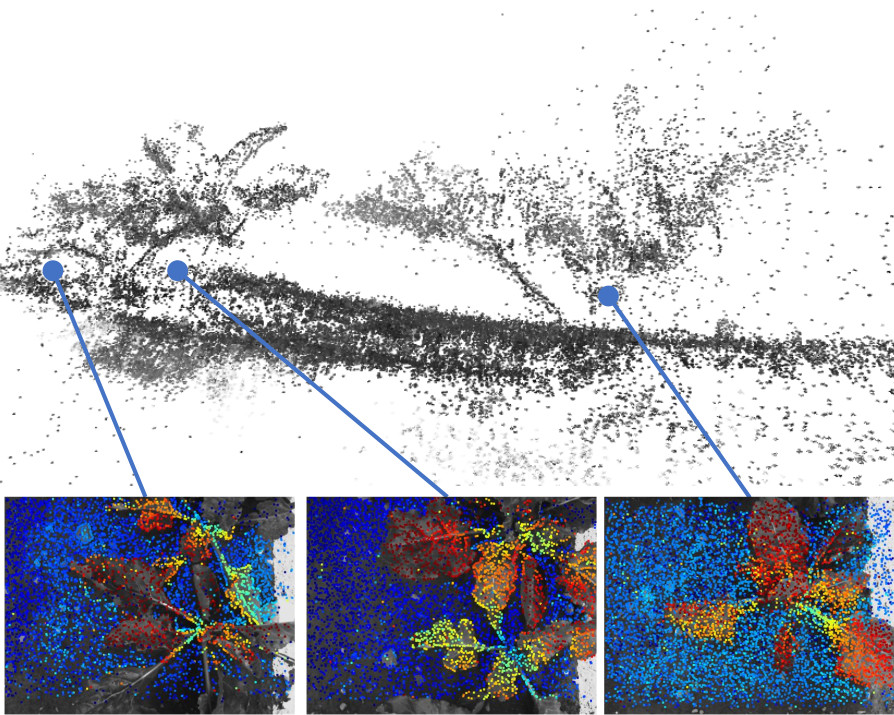}
   \caption{
  An example of reconstructed inverse depth map and a 3D point-cloud of plants and ground surface from our proposed intra-camera tracking algorithm.}
   \label{fig:trackingoverview_matching}
\end{figure}

The main challenge in the weed tracking  with non-overlapping multi-camera systems (\figref{fig:bonirob-weed-unit},~\secref{sec:weed_unit_sensors}) is to deal with the high variance delay between the instant when the image of the first camera is acquired and the one when a target is detected by the detection system. To address this issue, a novel tracking system has been developed. The inputs are the images and the coordinates of the targets given by the classifier (Sec.~\ref{sec:_crop_weed_detection}) in the images of the detection camera (see \figref{fig:bonirob-weed-unit}).
The outputs are the the trigger time and position for the actuators. The main steps are illustrated in \figref{fig:bonirob-weed-unit} (top):
\begin{enumerate}
    \item The \emph{intra-camera tracking} module estimates the camera pose and the 3D scene map using VO direct methods;
    \item After receiving delayed classification results and scene structures, the \emph{object initializer and updater} module creates the templates of the received objects, propagates their updated poses, and accumulates their labels;
    \item To prevent destroying a misclassified crop, a naive Bayes classifier (NBC) validates its classification based on the accumulated labels;
    \item Once a new weed object moves into the tracking camera FoV, inter-camera tracking performs illumination-robust direct tracking to find its new pose and creates a new template for intra-camera tracking;
    \item After repeated intra-camera tracking, updating, and inter-camera tracking, the weed finally approaches the endeffector, where the control algorithm predicts the trigger time and position of actuation for intervention.
\end{enumerate}

The novelty in this module resides in intra- and inter-camera tracking~\cite{wu2019design} (\figref{fig:trackingoverview_matching}).

\emph{Intra-camera tracking:} unlike conventional multi-object tracking algorithms, our proposed VO approach recovers the 3D scene structure before obtaining object information, then formulates each template as a combination of trimmed image and inverse depth map for later tracking upon arrival of classification results.
This strategy guarantees a constant-time operation despite the change of the amount of tracked objects. 

\emph{Inter-camera tracking:} taking advantage that only the 2D positions of weeds in image space is of interest, we extract the small frame template of each weed combined with a global illumination-invariant cost to perform local image alignment. Then, the weed center and its template boundary are transformed into the current frame using the pose estimate, which is used to generate a new template for intra-camera tracking.
To be robust to changes of viewpoint, the retrieval of weeds objects is achieved by using 3D-2D direct template-based matching.

\begin{figure}[t]
   \includegraphics[width=\linewidth]{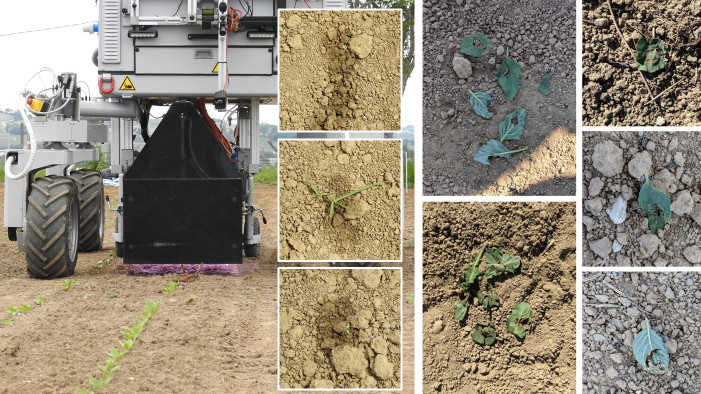}
   \center{
   \includegraphics[width=0.8\linewidth]{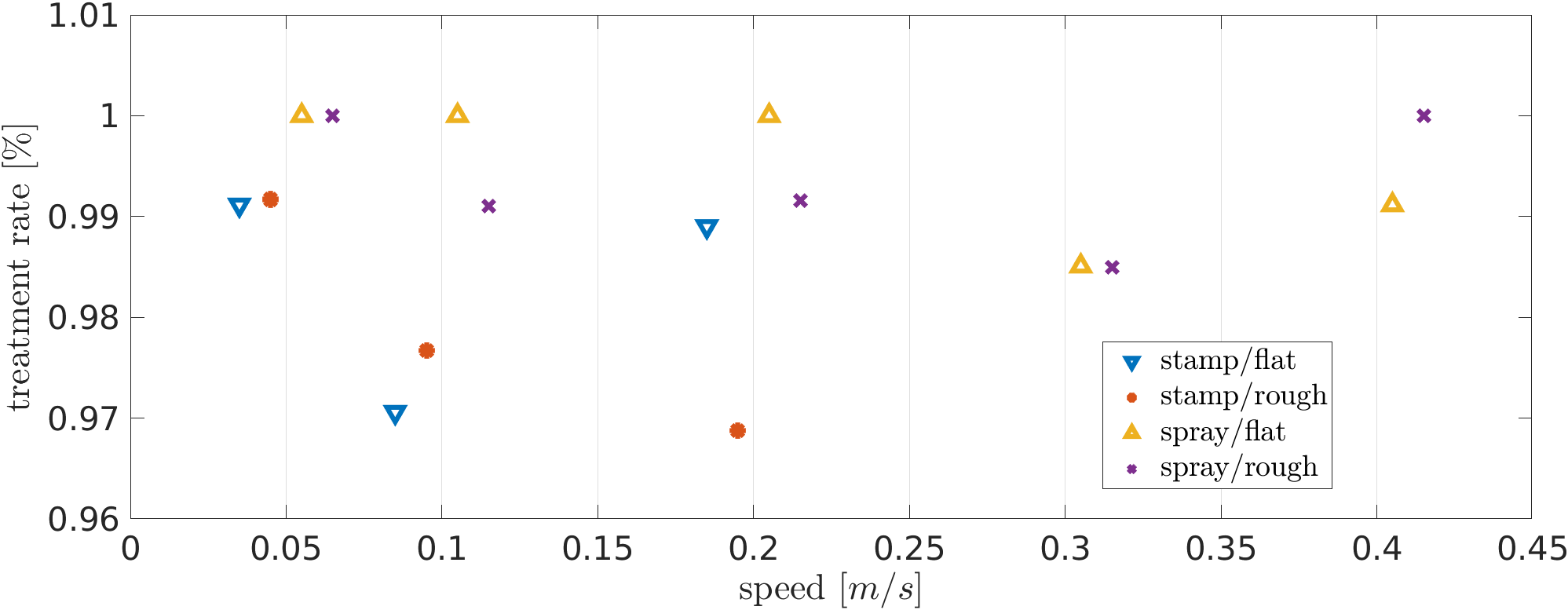}
   }
   \caption{Top: experiments in real environment for spraying evaluation, in real field with fake weeds for stamping evaluation.
   Bottom: treatment rates results both for stamping and spraying in rough and flat environment.}
   \label{fig:fieldxp}
\end{figure}

To evaluate the mechanical weed removal, real leaves with an average radius of 10mm are chosen as targets, counting the successfully stamped ones.
To evaluate selective spraying, we set up a webcam to monitor targets after spraying.
These experiments and the results are illustrated in \figref{fig:fieldxp}, in which
we can observe that the successful treatment rate is almost invariant with the speed in both flat and rough field ground.

\subsection{Agronomic Impacts in Sugar Beet Crops}\label{sec:agronomic_impact}

The potential impacts of the Flourish methodologies have been investigated through a 3-years field campaign carried out in Italy. We compared traditional full-field herbicide treatments with a chemical weeding system, targeting the areas with higher weeds density and therefore simulating the Flourish selective intervention. We performed four trials on sugar beet experimental plots grown under the same agronomic conditions and on the same site. For each year, four $10 m^2$ plots (one plot per trial) have been used; in 2017 and 2018 each trial was replicated three times.


For each trial, different pre-emergence treatments (PRT) and post-emergence treatments (POT) were performed before and after the seedling emergence, using dicotyledons/monocotyledons herbicides. \tabref{tab:sugar_production} reports the percentages of plot subject to treatment (PRT and POT columns).

At the end of each crop cycle, sugar beets roots of each plot were harvested and weighed. Representative samples from each plot were delivered to ASSAM laboratories for the refractometric estimation of the average root sucrose content (Brix degrees). The amount of sucrose produced in each experimental plot was then computed and then related to the agronomic surface unit (tons of sucrose per hectare). Results are reported in \tabref{tab:sugar_production}. The 2017 and 2018 results were calculated for each trial by averaging the results of the 3 replicates.

\begin{table}[htb]
\caption{Treatments and tons of sucrose per hectare produced in the 2016-2018 trials.}
\begin{center}
\begin{tabular}{|m{1.3cm}m{1cm}m{1cm}|m{1cm}m{1cm}m{1cm}|}
\hline \vspace{+1mm}
\textbf{Trials}	& PRT & POT & 2016 & 2017 & 2018
\\ \hline\vspace{+1mm}
\textbf{Trial A}	& 100\% & 100\% & 15.2	& 8.4	& 10.1
\\
\textbf{Trial B}	& 100\% & 30\%
& 12.3	& 8.5	& 11.5
\\
\textbf{Trial C}	& 100\% & 0 & 11.7	& 4.9	& 11.2
\\
\textbf{Trial D}	& 0 & 0 & 11.0	& 3.4	& 6.6
\\ \hline
\end{tabular}\label{tab:sugar_production}
\end{center}
\end{table}

Results suggest that sugar beet selective POT (Trial B), when associated to an ordinary PRT, represents a sustainable alternative to conventional POT full-field-treatments (Trial A), as it allows to achieve comparable sucrose production levels while reducing chemical inputs. In fact, compared to Trial A the average sucrose production of Trial B was just 3.9\% lower over the three years. At the same time both selective and full-field POT turned out to be an effective support to production, especially in the case of crops subjected to marked grass-weeds pressure. Traditional full-field PRT was essential in controlling weeds (mainly dicotyledons) in a slow-growing sugar beet crop (Trial D average production 24.5\% lower than Trial C).


Besides the impacts on production and environment, the introduction of precision agriculture technologies into ordinary cultural practices could also have a larger-scale effect on the farming sector. To support this intuition, a participatory evaluation (Metaplan model and SWOT analysis) carried out within the project and involving a panel of 17 stakeholders professionally operating in the farming sector, highlighted that the application of such technologies would be able to potentially improve the efficiency, efficacy and safety of farming operations, to reduce the labor cost, to provide more information on crops, field structure, and meteorological events, to increase farming environmental sustainability and food safety. 

\section{Open-Source Software and Datasets}\label{sec:datasets}

Many of the methods presented above have been released as open-source software, with download links reported in the corresponding papers. A short-list is given here:
\begin{itemize}
    \item A modified version of DJI Onboard ROS Software Development Kit (SDK)~\cite{Sa2017};
    \item Plant stress phenotyping dataset and analysis software~\cite{Khanna2019} (\secref{sec:uav_localization}).
    \item The IPP framework (\secref{sec:uav_planning}) for terrain monitoring ~\cite{Popovic2020}\footnote{\url{https://github.com/ethz-asl/waypoint_navigator}}; 
    \item MCAPS (\secref{sec:ugv_localization})~\cite{ipngp_RA-L2018};
    \item AgriColMap (\secref{sec:cooperative_modeling})~\cite{pknsnp_RA-L2019};
    \item Algorithms for synchronizing clocks\footnote{\url{https://github.com/ethz-asl/cuckoo_time_translator}}.
\end{itemize}


We also created and made publicly available several novel datasets:


\begin{itemize}
    \item Sugar Beets 2016: a novel, vast long-term dataset of a sugar beet field~\cite{chebrolu2017ijrr}. 
    \item Flourish Sapienza Datasets \cite{flourish_sapeinza_datasets}: a collection of datasets, with related ground truths, acquired from farming robots;
    \item A dataset of 4-, 5-channel multi-spectral aerial images dedicated to plant semantic segmentation~\cite{sa2018ral,sa2018rs}; 
    \item A pixel-wise ground truthed sugar beet and weed datasets collected from a controlled field experiment \cite{sa2018ral}\footnote{\url{https://goo.gl/UK2pZq}}; 
    \item WeedMap dataset~\cite{sa2018rs}, contains high-fidelity, large-scale, and spatio-temporal multi-spectral images.
\end{itemize}



\section{Conclusions}

The main goal of the Flourish research project was to develop an adaptable robotic solution for precision farming by combining the aerial survey capabilities of a small autonomous UAV with a multi-purpose agricultural UGV. 
In this paper, we presented an overview of the custom-built hardware solutions, methods and algorithms developed in the project, which are tailored for cooperation between aerial and ground robots. We demonstrate a successful in-field intervention task integrating the various modules.

We believe that the proposed solutions represent, from several points of view, a step forward in the state-of-the-art of robotic systems applied to precision agriculture, with solutions that are easily applicable to a wide range of robots, farm management activities, and crop types.

\bibliographystyle{IEEEtran}
\balance
\bibliography{flourish_overview}

\begin{thebibliography}{10}
\providecommand{\url}[1]{#1}
\csname url@rmstyle\endcsname
\providecommand{\newblock}{\relax}
\providecommand{\bibinfo}[2]{#2}
\providecommand\BIBentrySTDinterwordspacing{\spaceskip=0pt\relax}
\providecommand\BIBentryALTinterwordstretchfactor{4}
\providecommand\BIBentryALTinterwordspacing{\spaceskip=\fontdimen2\font plus
\BIBentryALTinterwordstretchfactor\fontdimen3\font minus
  \fontdimen4\font\relax}
\providecommand\BIBforeignlanguage[2]{{%
\expandafter\ifx\csname l@#1\endcsname\relax
\typeout{** WARNING: IEEEtran.bst: No hyphenation pattern has been}%
\typeout{** loaded for the language `#1'. Using the pattern for}%
\typeout{** the default language instead.}%
\else
\language=\csname l@#1\endcsname
\fi
#2}}

\bibitem{flourish_project}
\BIBentryALTinterwordspacing
Flourish project. [Online]. Available: \url{http://flourish-project.eu}
\BIBentrySTDinterwordspacing

\bibitem{rhea_project}
\BIBentryALTinterwordspacing
Rhea project. [Online]. Available: \url{http://www.rhea-project.eu/}
\BIBentrySTDinterwordspacing

\bibitem{gasparri_h2020_2018}
A.~Gasparri, G.~Ulivi, N.~Bono~Rossello, and E.~Garone, ``The {H}2020 project
  {Pantheon}: precision farming of hazelnut orchards,'' in \emph{Convegno
  Automatica.it}, 2018.

\bibitem{astofli_grapes_2018}
P.~Astolfi, A.~Gabrielli, L.~Bascetta, and M.~Matteucci, ``Vineyard autonomous
  navigation in the echord++ grape experiment,'' \emph{IFAC-PapersOnLine},
  vol.~51, pp. 704--709, 01 2018.

\bibitem{sweeper_project}
\BIBentryALTinterwordspacing
Sweeper project. [Online]. Available: \url{http://www.sweeper-robot.eu/}
\BIBentrySTDinterwordspacing

\bibitem{Dong2019}
W.~Dong, P.~Roy, and V.~Isler, ``Semantic mapping for orchard environments by
  merging two-sides reconstructions of tree rows,'' \emph{Journal of Field
  Robotics}, vol.~37, no.~1, pp. 97--121, 2020.

\bibitem{Chen2017}
S.~W. {Chen}, S.~S. {Shivakumar}, S.~{Dcunha}, J.~{Das}, E.~{Okon}, C.~{Qu},
  C.~J. {Taylor}, and V.~{Kumar}, ``Counting apples and oranges with deep
  learning: A data-driven approach,'' \emph{IEEE Robotics and Automation
  Letters}, vol.~2, no.~2, pp. 781--788, April 2017.

\bibitem{Ehsani2016}
R.~Ehsani, D.~Wulfsohn, J.~Das, I.~Lagos, and Z.~Inés, ``Yield estimation; a
  low-hanging fruit for application of small uas.'' \emph{Resource: Engineering
  and Technology for Sustainable World}, vol.~23, pp. 16--18, 2016.

\bibitem{Das2015}
J.~{Das}, G.~{Cross}, C.~{Qu}, A.~{Makineni}, P.~{Tokekar}, Y.~{Mulgaonkar},
  and V.~{Kumar}, ``Devices, systems, and methods for automated monitoring
  enabling precision agriculture,'' in \emph{Proc. of the IEEE International
  Conference on Automation Science and Engineering (CASE)}, 2015, pp. 462--469.

\bibitem{ecorobotix}
\BIBentryALTinterwordspacing
{E}co{R}obotix. [Online]. Available: \url{https://www.ecorobotix.com/en/}
\BIBentrySTDinterwordspacing

\bibitem{blueriver}
\BIBentryALTinterwordspacing
Blueriver technology. [Online]. Available:
  \url{http://about.bluerivertechnology.com/}
\BIBentrySTDinterwordspacing

\bibitem{sagarobotics}
\BIBentryALTinterwordspacing
Saga~robotics. [Online]. Available: \url{https://sagarobotics.com/}
\BIBentrySTDinterwordspacing

\bibitem{Nikolic2014}
J.~Nikolic, J.~Rehder, M.~Burri, P.~Gohl, S.~Leutenegger, P.~T. Furgale, and
  R.~Siegwart, ``{A synchronized visual-inertial sensor system with FPGA
  pre-processing for accurate real-time SLAM},'' in \emph{Proc.~of the IEEE
  International Conference on Robotics and Automation (ICRA)}, 2014, pp.
  431--437.

\bibitem{lottes2017jfr}
P.~Lottes, M.~H\"oferlin, S.~Sander, and C.~Stachniss, ``{Effective
  Vision-based Classification for Separating Sugar Beets and Weeds for
  Precision Farming},'' \emph{Journal of Field Robotics}, vol.~34, pp.
  1160--1178, 2017.

\bibitem{lottes2017icra}
P.~Lottes, R.~Khanna, J.~Pfeifer, R.~Siegwart, and C.~Stachniss, ``{UAV-based
  Crop and Weed Classification for Smart Farming},'' in \emph{Proc.~of the IEEE
  International Conference on Robotics and Automation (ICRA)}, 2017.

\bibitem{lottes2018ral}
P.~Lottes, J.~Behley, A.~Milioto, and C.~Stachniss, ``Fully convolutional
  networks with sequential information for robust crop and weed detection in
  precision farming,'' \emph{IEEE Robotics and Automation Letters}, vol.~3,
  no.~4, pp. 3097--3104, 2018.

\bibitem{lottes2017iros}
P.~Lottes and C.~Stachniss, ``Semi-supervised online visual crop and weed
  classification in precision farming exploiting plant arrangement,'' in
  \emph{Proc. of the IEEE/RSJ International Conference on Intelligent Robots
  and Systems (IROS)}, 2017.

\bibitem{sa2018ral}
I.~Sa, Z.~Chen, M.~Popovi{\'{c}}, R.~Khanna, F.~Liebisch, J.~Nieto, and
  R.~Siegwart, ``{weedNet: Dense Semantic Weed Classification Using
  Multispectral Images and MAV for Smart Farming},'' \emph{IEEE Robotics and
  Automation Letters}, vol.~3, no.~1, pp. 588--595, 2018.

\bibitem{dpgp_IROS2017}
M.~Di~Cicco, C.~Potena, G.~Grisetti, and A.~Pretto, ``Automatic model based
  dataset generation for fast and accurate crop and weeds detection,'' in
  \emph{Proc. of the IEEE/RSJ International Conference on Intelligent Robots
  and Systems (IROS)}, 2017.

\bibitem{Hofmann_2005}
C.~M. Hoffmann, ``Changes in nitrogen composition of sugar beet varieties in
  response to increasing nitrogen supply,'' \emph{Journal of Agronomy and Crop
  Science}, vol. 191, no.~2, pp. 138--145, 2005.

\bibitem{Walter_ISPA_2018}
A.~Walter, R.~Khanna, P.~Lottes, C.~Stachniss, R.~Siegwart, J.~Nieto, and
  F.~Liebisch, ``A robotic approach for automation in crop management,'' in
  \emph{Proc. of the International Conference on Precision Agriculture}, 2018.

\bibitem{Walter6148}
A.~Walter, R.~Finger, R.~Huber, and N.~Buchmann, ``Opinion: Smart farming is
  key to developing sustainable agriculture,'' \emph{Proceedings of the
  National Academy of Sciences}, vol. 114, no.~24, pp. 6148--6150, 2017.

\bibitem{Finger_2019}
R.~Finger, S.~M. Swinton, N.~E. Benni, and A.~Walter, ``Precision farming at
  the nexus of agricultural production and the environment,'' \emph{Annual
  Review of Resource Economics}, vol.~11, no.~1, 2019.

\bibitem{Argento_2019_ECPA}
F.~Argento, T.~Anken, F.~Liebisch, and A.~Walter, ``Crop imaging and soil
  adjusted variable rate nitrogen application in winter wheat,'' in
  \emph{Precision agriculture ’19}, 2019, pp. 511 -- 517.

\bibitem{Bloesch2015}
M.~Bloesch, S.~Omari, M.~Hutter, and R.~Siegwart, ``Robust visual inertial
  odometry using a direct ekf-based approach,'' in \emph{Proc. of the IEEE/RSJ
  International Conference on Intelligent Robots and Systems (IROS)}, 2015, pp.
  298--304.

\bibitem{Lynen2013}
S.~Lynen, M.~Achtelik, S.~Weiss, M.~Chli, and R.~Siegwart, ``A robust and
  modular multi-sensor fusion approach applied to mav navigation,'' in
  \emph{Proc. of the IEEE/RSJ International Conference on Intelligent Robots
  and Systems (IROS)}, 2013.

\bibitem{Schneider2018}
T.~Schneider, M.~Dymczyk, M.~Fehr, K.~Egger, S.~Lynen, I.~Gilitschenski, and
  R.~Siegwart, ``maplab: An open framework for research in visual-inertial
  mapping and localization,'' \emph{arXiv preprint 1711.10250}, 2018.

\bibitem{ipngp_RA-L2018}
M.~Imperoli, C.~Potena, D.~Nardi, G.~Grisetti, and A.~Pretto, ``An effective
  multi-cue positioning system for agricultural robotics,'' \emph{IEEE Robotics
  and Automation Letters}, vol.~3, no.~4, pp. 3685--3692, October 2018.

\bibitem{pknsnp_RA-L2019}
C.~Potena, R.~Khanna, J.~Nieto, R.~Siegwart, D.~Nardi, and A.~Pretto,
  ``{A}gri{C}ol{M}ap: {A}erial-ground collaborative {3D} mapping for precision
  farming,'' \emph{IEEE Robotics and Automation Letters}, vol.~4, no.~2, pp.
  1085--1092, 2019.

\bibitem{hu2016}
Y.~Hu, R.~Song, and Y.~Li, ``Efficient coarse-to-fine patch match for large
  displacement optical flow,'' in \emph{Proc.~of the Conference on Computer
  Vision and Pattern Recognition (CVPR)}, 2016, pp. 5704--5712.

\bibitem{myronenko2010}
A.~Myronenko and X.~Song, ``Point set registration: Coherent point drift,''
  \emph{IEEE Transactions on Pattern Analysis and Machine Intelligence},
  vol.~32, no.~12, pp. 2262--2275, 2010.

\bibitem{chebrolu2018ral}
N.~Chebrolu, T.~L\"abe, and C.~Stachniss, ``{Robust Long-Term Registration of
  UAV Images of Crop Fields for Precision Agriculture},'' \emph{IEEE Robotics
  and Automation Letters}, vol.~3, no.~4, pp. 3097--3104, 2018.

\bibitem{Popovic2020}
M.~Popovi{\'{c}}, T.~Vidal-Calleja, G.~Hitz, J.~J. Chung, I.~Sa, R.~Siegwart,
  and J.~Nieto, ``{An informative path planning framework for UAV-based terrain
  monitoring},'' \emph{Autonomous Robots}, vol.~44, pp. 889--911, 2020.

\bibitem{winterhalter18ral}
W.~Winterhalter, F.~Fleckenstein, C.~Dornhege, and W.~Burgard, ``{Crop Row
  Detection on Tiny Plants With the Pattern Hough Transform},'' \emph{IEEE
  Robotics and Automation Letters}, vol.~3, no.~4, pp. 3394--3401, 2018.

\bibitem{fleckenstein17icra}
F.~Fleckenstein, C.~Dornhege, and W.~Burgard, ``{Efficient Path Planning for
  Mobile Robots with Adjustable Wheel Positions},'' in \emph{Proc.~of the IEEE
  International Conference on Robotics and Automation (ICRA)}, 2017.

\bibitem{fleckenstein19fsr}
F.~Fleckenstein, W.~Winterhalter, C.~Dornhege, C.~Pradalier, and W.~Burgard,
  ``{Smooth Local Planning Incorporating Steering Constraints},'' in
  \emph{Proc. of the 12th Conference on Field and Service Robotics (FSR)},
  2019.

\bibitem{pradalier:hal-01435823}
\BIBentryALTinterwordspacing
C.~Pradalier, ``{A task scheduler for ROS},'' Jan. 2017. [Online]. Available:
  \url{https://hal.archives-ouvertes.fr/hal-01435823}
\BIBentrySTDinterwordspacing

\bibitem{wu2019design}
X.~Wu, S.~Aravecchia, and C.~Pradalier, ``Design and implementation of computer
  vision based in-row weeding system,'' in \emph{Proc.~of the IEEE
  International Conference on Robotics and Automation (ICRA)}.\hskip 1em plus
  0.5em minus 0.4em\relax IEEE, 2019, pp. 4218--4224.

\bibitem{Sa2017}
I.~Sa, M.~Kamel, M.~Burri, M.~Bloesch, R.~Khanna, M.~Popovi{\'{c}}, J.~Nieto,
  and R.~Siegwart, ``Build your own visual-inertial drone: A cost-effective and
  open-source autonomous drone,'' \emph{IEEE Robotics \& Automation Magazine},
  vol.~25, no.~1, pp. 89--103, 2018.

\bibitem{Khanna2019}
R.~Khanna, L.~Schmid, A.~Walter, J.~Nieto, R.~Siegwart, and F.~Liebisch, ``A
  spatio temporal spectral framework for plant stress phenotyping,''
  \emph{Plant Methods}, vol.~15, 2019.

\bibitem{chebrolu2017ijrr}
N.~Chebrolu, P.~Lottes, A.~Schaefer, W.~Winterhalter, W.~Burgard, and
  C.~Stachniss, ``Agricultural robot dataset for plant classification,
  localization and mapping on sugar beet fields,'' \emph{Journal of Robotics
  Research}, vol.~36, no.~10, pp. 1045--1052, 2017.

\bibitem{flourish_sapeinza_datasets}
``{F}lourish {S}apienza {D}atasets~~~[{O}nline],'' in
  \emph{https://www.dis.uniroma1.it/\texttildelow labrococo/fsd/}.

\bibitem{sa2018rs}
I.~Sa, M.~Popovi{\'{c}}, R.~Khanna, Z.~Chen, P.~Lottes, F.~Liebisch, J.~Nieto,
  C.~Stachniss, A.~Walter, and R.~Siegwart, ``{WeedMap: A Large-Scale Semantic
  Weed Mapping Framework Using Aerial Multispectral Imaging and Deep Neural
  Network for Precision Farming},'' \emph{Remote Sensing}, vol.~10, no.~9,
  2018.

\end{thebibliography}
\end{document}